\documentclass{article}

\usepackage{mathrsfs}
\usepackage{amssymb}
\usepackage{mathtools}
\usepackage{amsmath,amsthm,amsfonts}
\usepackage{enumerate}   
\usepackage{microtype}
\usepackage{graphicx}
\usepackage{subfigure}
\usepackage{booktabs} 
\usepackage{hyperref}
\usepackage{tikz}
\usepackage[capitalize,noabbrev]{cleveref}
\usepackage[skip=0pt]{caption}
\usepackage{subcaption}

\usepackage{autonum}
\usepackage{bbm}
\usepackage{bm}
\usepackage{csquotes}
\usepackage{multirow}
\usepackage{enumitem}
\usepackage{pifont}

\usepackage[accepted]{icml2024}

\newcommand*\circled[1]{\tikz[baseline=(char.base)]{
            \node[shape=circle,draw,inner sep=1.pt,fill=black] (char) {\textcolor{white}{\textbf{#1}}};}}

\DeclareMathOperator*{\argmin}{arg\,min}

\newcommand*{\N}{\mathbb{N}}
\DeclareMathOperator*{\E}{\mathbbm{E}}

\DeclareMathOperator{\R}{\mathbb{R}}

\RequirePackage{algpseudocode}
\renewcommand{\algorithmicrequire}{\textbf{Input:}}
\renewcommand{\algorithmicensure}{\textbf{Output:}}

\makeatletter
\renewcommand\paragraph{\@startsection{paragraph}{4}{\z@}%
                                    {0ex \@plus.0ex \@minus.0ex}%
                                    {-1em}%
                                    {\normalfont\normalsize\bfseries}}

\theoremstyle{plain}
\newtheorem{theorem}{Theorem}[section]

\theoremstyle{definition}

\theoremstyle{remark}
\newtheorem{remark}[theorem]{Remark}

\icmltitlerunning{Solving Poisson Equations using Neural Walk-on-Spheres}

\makeatother

\begin{document}

\twocolumn[
\icmltitle{Solving Poisson Equations using Neural Walk-on-Spheres}

\icmlsetsymbol{equal}{*}

\begin{icmlauthorlist}
\icmlauthor{Hong Chul Nam}{equal,eth}
\icmlauthor{Julius Berner}{equal,caltech}
\icmlauthor{Anima Anandkumar}{caltech}
\end{icmlauthorlist}

\icmlaffiliation{eth}{ETH Zurich}
\icmlaffiliation{caltech}{Caltech}
\icmlcorrespondingauthor{Hong Chul Nam}{\href{mailto:honam@student.ethz.ch}{honam@student.ethz.ch}}
\icmlcorrespondingauthor{Julius Berner}{\href{mailto:jberner@caltech.edu}{jberner@caltech.edu}}

\icmlkeywords{Machine Learning, ICML}

\vskip 0.3in
]



\printAffiliationsAndNotice{\icmlEqualContribution} 

\begin{abstract}
We propose \emph{Neural Walk-on-Spheres} (NWoS), a novel neural PDE solver for the efficient solution of high-dimensional Poisson equations. Leveraging stochastic representations and Walk-on-Spheres methods, we develop novel losses for neural networks based on the recursive solution of Poisson equations on spheres inside the domain. The resulting method is highly parallelizable and does not require spatial gradients for the loss. We provide a comprehensive comparison against competing methods based on PINNs, the Deep Ritz method, and (backward) stochastic differential equations. In several challenging, high-dimensional numerical examples, we demonstrate the superiority of NWoS in accuracy, speed, and computational costs. Compared to commonly used PINNs, our approach can reduce memory usage and errors by orders of magnitude. Furthermore, we apply NWoS to problems in PDE-constrained optimization and molecular dynamics to show its efficiency in practical applications.
\end{abstract}

\section{Introduction}

Partial Differential Equations (PDE) are foundational to our modern scientific understanding in a wide range of domains. While decades of research have been devoted to this topic, numerical methods to solve PDEs remain expensive for many PDEs. In recent years, deep learning has helped to accelerate the solution of PDEs~\cite{azzizadenesheli2023neural,zhang2023artificial,cuomo2022scientific} as well as tackle PDEs, which had been entirely out of range for classical methods~\cite{han2018solving,scherbela2022solving,nusken2021solving}. 

Among the biggest challenges for classical numerical PDE solvers are complex geometries and high dimensions. In particular,  grid-based methods, such as finite-element, finite-volume, or finite-difference methods, scale exponentially in the underlying dimension. On the other hand, deep learning approaches have been shown to overcome this so-called \emph{curse of dimensionality}~\cite{de2022error,duan2021convergence,berner2020analysis}. 
Corresponding algorithms are typically based on \emph{Monte Carlo} (MC) approximations of variational formulations of PDEs.

In this work, we focus on high-dimensional Poisson equations on general domains. We note that the accurate numerical solution of such types of PDEs is crucial for a large variety of areas. For instance, Poisson equations are prominent in geometry processing~\cite{sawhney2020monte}, as well as many areas of theoretical physics, e.g., electrostatics and
quantum mechanics~\cite{bahrami2014schrodinger}. In high dimensions, they govern 
important quantities in molecular dynamics, such as likely transition pathways and transition rates between regions or conformations of interest~\cite{vanden2006towards,lu2015reactive}.

Several deep learning methods are amendable to the numerical solution of Poisson equations. This includes physics-informed neural networks (PINNs) or Deep Galerkin methods~\cite{raissi2019physics,sirignano2018dgm}, the Deep Ritz method~\cite{weinan2017deep}, as well as approaches based on (backward) stochastic differential equations~\cite{han2017deep,nusken2021interpolating,han2020derivative}. However, previous methods suffer from unnecessarily high computational costs, bias, or instabilities, see Section~\ref{sec:methods}.

\begin{figure}[t]
    \centering
    \includegraphics[width=0.85\linewidth]{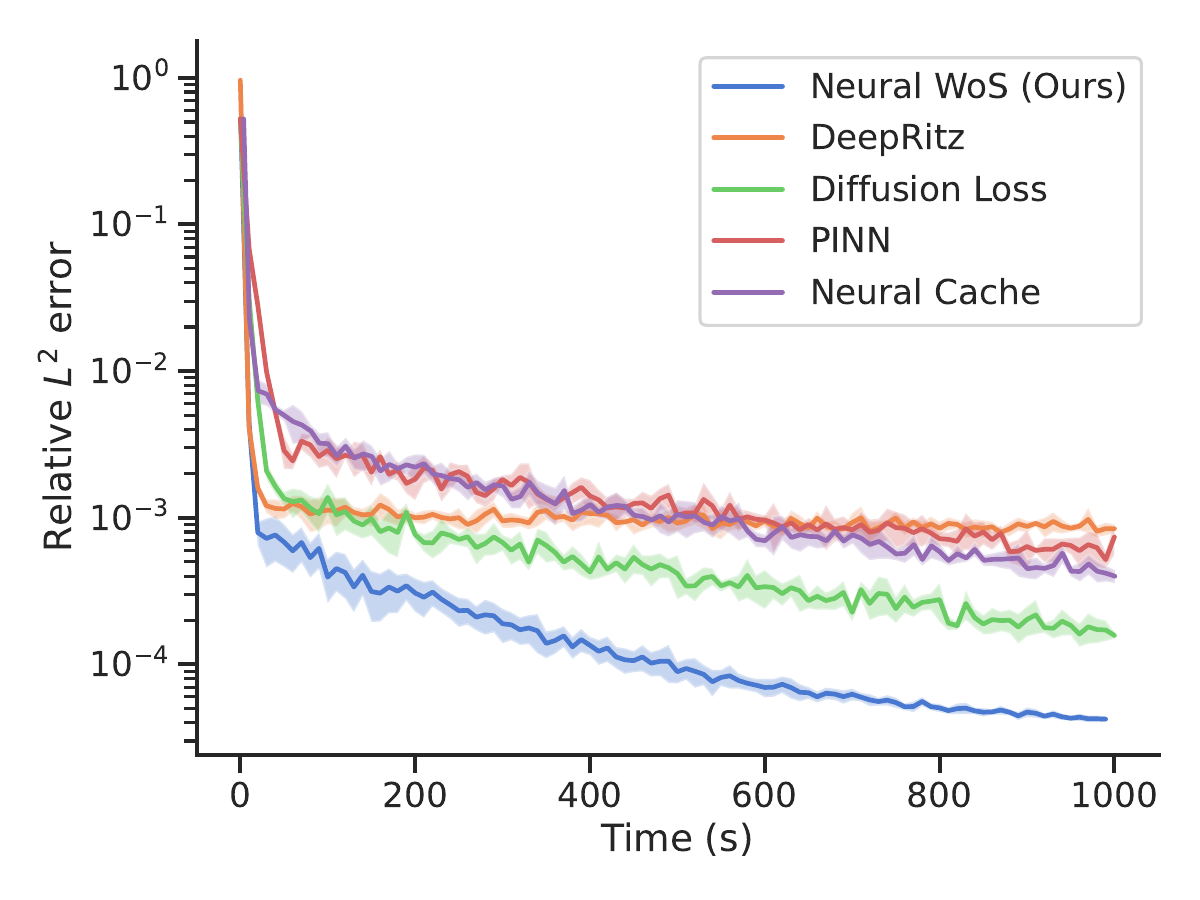}
    \vspace{-0.75em}
    \caption{Convergence of the relative $L^2$-error when solving the 10$d$  Laplace equation in~\Cref{sec:numerics} using our considered methods.}
    \label{fig:convergence}
    \vspace{-2.0em}
\end{figure}

\begin{figure}[t]
    \centering
    \begin{minipage}{0.5\linewidth}
        \centering
        \includegraphics[width=0.8\linewidth]{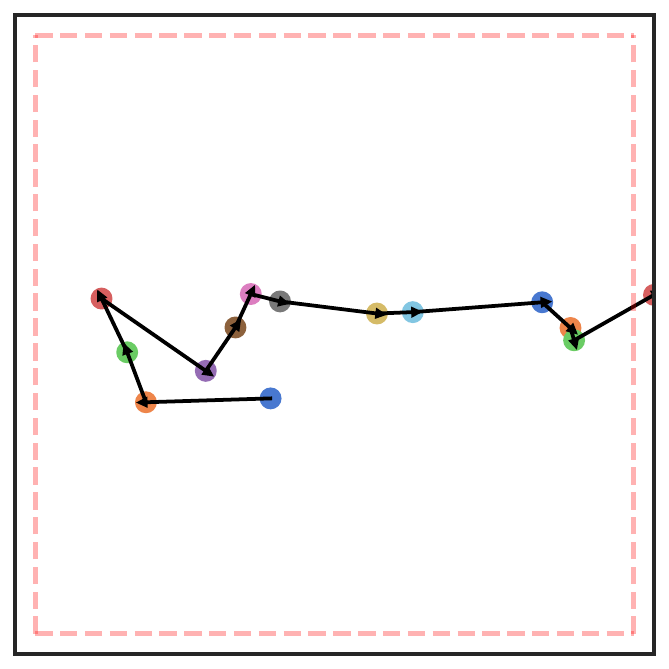}
    \end{minipage}%
    \begin{minipage}{0.5\linewidth}
        \centering
        \includegraphics[width=0.8\linewidth]{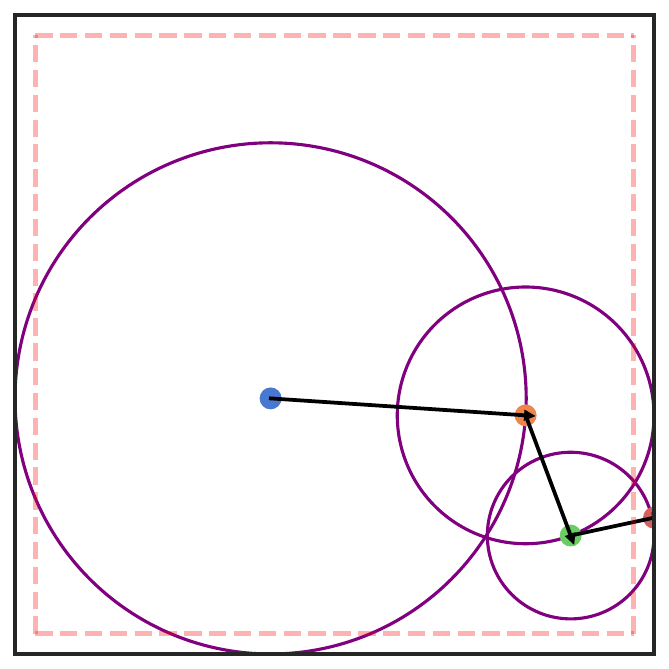}
    \end{minipage}
    \caption{\textbf{Left:} Time-discretization of the solution $X^\xi$ to the SDE in~\eqref{eq:sde} with stopping time $\tau(\Omega, \xi)$ in~\eqref{eq:stopping_time} for the domain $\Omega =[0,1]^2$. \\ \textbf{Right:} Realization of the Walk-on-Spheres algorithm in~\Cref{sec:nwos}.} 
    \label{fig:wos}
    \vspace{-0.5em}
\end{figure}

\paragraph{Walk-on-Spheres (WoS):} To overcome the above challenges, we propose a novel 
approach based on so-called \emph{Walk-on-Spheres} (WoS) methods~\cite{muller1956some}. The WoS method is a Monte Carlo method specifically tailored toward Poisson equations by rewriting their solutions as 
an expectation over Brownian motions stopped at the boundary of the domain. Simulating the Brownian motion using time-discretizations either is slow or introduces bias (depending on the chosen time step). Leveraging the isotropy of Brownian motion, WoS accelerates this process by iteratively sampling from spheres around the current position until reaching the boundary, see \Cref{fig:wos}. 

However, as with all Monte Carlo methods, WoS can only obtain \emph{pointwise} estimates and suffers from slow convergence w.r.t.\@ the number of trajectories. 
In particular, every sufficiently accurate estimate of the solution on a single point takes a considerable amount of time. This is prohibitive if many solution evaluations are needed sequentially, e.g., in PDE-constrained optimization problems.  

\paragraph{Our approach (NWoS):} We develop \emph{Neural Walk-on-Spheres} (NWoS), a version of WoS that can be combined with neural networks to learn the solution to (parametric families of) Poisson equations on the \emph{whole domain}. Our method amortizes the cost of WoS during training so that the solution, and its gradients, can be evaluated in fractions of seconds afterward (and at arbitrary points in the domain). 
In particular, in order to obtain accuracy $\varepsilon$, the standard WoS method incurs a cost of  $\mathcal{O}(\varepsilon^{-2})$ trajectories for the evaluation of the solution while NWoS has a reduced cost of a single $\mathcal{O}(1)$ forward-pass of our model. 

\begin{figure}[t]
    \centering
    \includegraphics[width=0.8\linewidth]{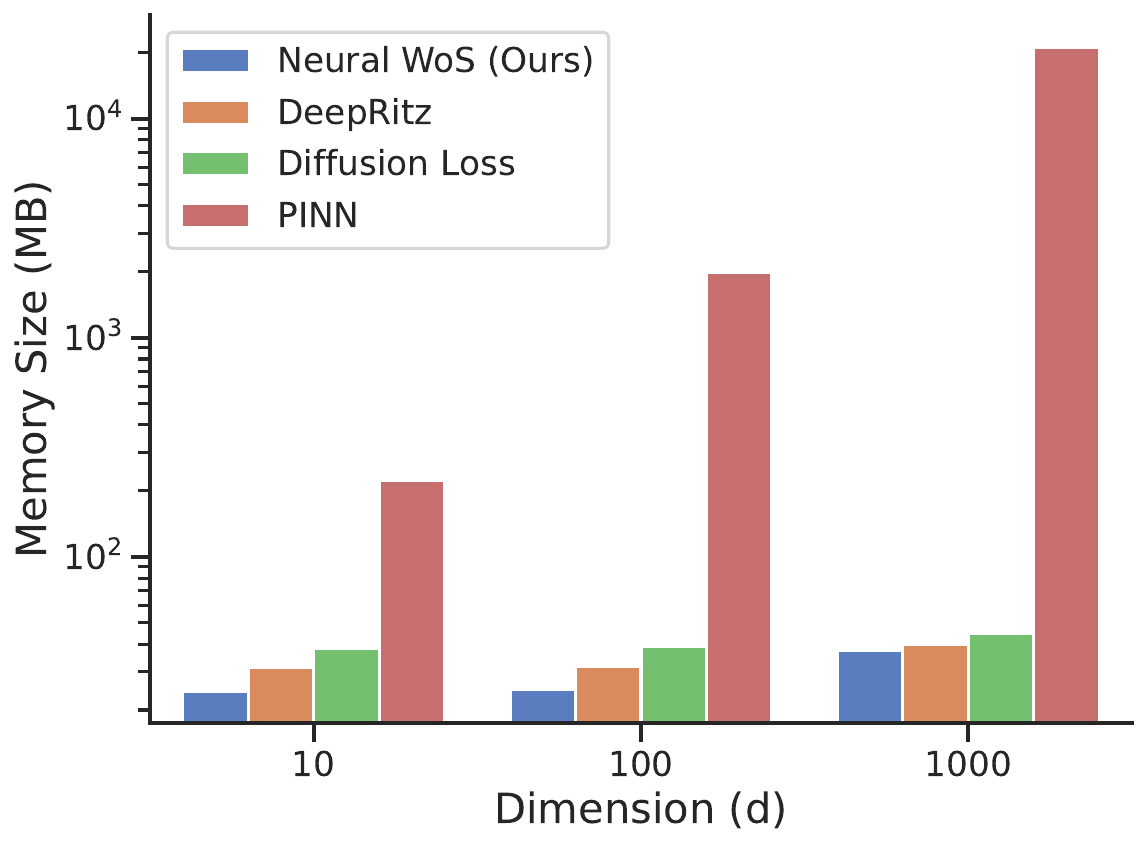}
    \vspace{-0.8em}
    \caption{Peak GPU memory usage of different methods during training with batch size $512$ for the Poisson equation in \Cref{sec:numerics} in different dimensions $d$.}
    \label{fig:mem}
    \vspace{-0.5em}
\end{figure}

Using the partially trained model as an estimator, we can limit the number of simulations and WoS steps for training without introducing high bias or variance. 
The resulting objective is more efficient and scalable than competing methods, without the need to balance penalty terms for the boundary condition or compute spatial derivatives (\Cref{tab:compare}).  
In particular, we demonstrate a significant reduction of GPU memory usage in comparison to PINNs (\Cref{fig:mem}) and up to orders of magnitude better performance for a given time and compute budget (\Cref{fig:convergence}).

Our contributions can be summarized as follows:
\begin{itemize}[leftmargin=*,itemsep=1pt,topsep=0pt]
\item We analyze previous neural PDE solver and their shortcomings when applied to high-dimensional elliptic PDEs, such as Poisson equations (\Cref{sec:methods}).
\item We devise novel variational formulations for the solution of Poisson equations based on WoS methods and provide corresponding theoretical guarantees and efficient implementations (\Cref{sec:nwos}).
\item We compare 
against previous approaches on a series of benchmarks and demonstrate significant improvements in terms of accuracy, speed, and scalability (\Cref{sec:numerics}). 
\end{itemize}

\begin{table}[t]
\vskip -6pt
\caption{Comparison of neural PDE solver for Poisson equations. \emph{\#Derivatives}, \emph{\#Loss terms}, and \emph{Cost} denote the order of spatial derivatives, the number of terms required in the loss function, and the computational cost for one gradient step. \emph{Propagation speed} describes how quickly boundary information can propagate to the interior of the domain, see~\Cref{sec:methods} for details.}
\vspace{0.5em}
\centering
\small
\begin{tabular}{l@{\hspace{0.75em}}c@{\hspace{0.75em}}c@{\hspace{0.75em}}c@{\hspace{0.75em}}c}
\toprule
Method & \#Derivatives   & \#Loss & Cost & Propagation \\ 
 & & terms & & speed \\ \midrule
PINN & $2$ & $2$ & medium & slow  \\
Deep Ritz & $1$ & $2$ & low & slow \\
Feynman-Kac & $0$ & $1$ & high & fast  \\
BSDE & $1$ & $1$ & high & fast \\
Diffusion loss & $1$ & $2$ & medium & medium  \\ \midrule
\textbf{NWoS (ours)}\footnotemark & $\mathbf{0}$ & $\mathbf{1}$ & \textbf{low} & \textbf{fast}  \\
 \bottomrule
\end{tabular}
\label{tab:compare}
\vspace{-0.45em}
\end{table}
\footnotetext{While not necessary for NWoS, we note that the gradient of the model and an additional boundary loss can still be used to improve performance, see~\Cref{sec:implementation}.}

\section{Related works}

\paragraph{Neural PDE solver:} We provide an in-depth comparison to competing deep learning approaches to solve elliptic PDEs in Section~\ref{sec:methods}. These include physics-informed neural networks (PINNs)~\cite{raissi2019physics, sirignano2018dgm}, the Deep Ritz method~\citep{jin2017deep}, and the diffusion loss~\cite{nusken2021interpolating}, see also \Cref{tab:compare}. The diffusion loss can be viewed as an interpolation between PINNs and losses based on backward SDEs (BSDEs)~\cite{han2017deep,weinan2017deep,beck2019machine}.
Methods based on BSDEs and the Feynman-Kac formula~\cite{beck2018solving,berner2020numerically,richter2022robust} have been investigated for the solution of parabolic PDEs, where the SDE is stopped at a given terminal time. Due to costly simulation times, they cannot be applied efficiently to elliptic problems.
To combat this issue, we draw inspiration from Walk-on-Spheres methods. 

\paragraph{Monte Carlo (MC) methods:} Since grid-based methods cannot tackle high-dimensional PDEs, MC methods are typically used. For Poisson equations, the WoS method has been developed by~\citet{muller1956some} and has since been successfully used in various scientific settings~\cite{sabelfeld2017random,juba2016acceleration,bossy2010probabilistic} as well as recently in computer graphics~\cite{qi2022bidirectional,sawhney2022grid}. 
In the latter domain, caches based on boundary values~\cite{miller2023boundary} and neural networks~\cite{li2023neural} have been proposed to estimate the PDE solution across the domain and accelerate convergence. Our objective can be scaled to high-dimensional parametric PDEs and guarantees that its minimizer approximates the solution on the whole domain. We refer to~\citet{hermann2020deep,beznea2022monte} for related neural network approximation results.

\section{Neural PDE Solver for Elliptic PDEs}
\label{sec:methods}
We start by defining our problem and describing previous deep learning methods for its solution. Our goal is to approximate the solution\footnote{For simplicity, we assume that a sufficiently smooth, strong solution exists.} $u\in C(\Omega)$ to elliptic PDEs with Dirichlet boundary conditions of the form
\begin{equation}
\label{eq:pde}
    \begin{cases}
    \mathcal{P}[u] = f, \quad &\text{on} \quad \Omega, \\
    u = g, &\text{on} \quad \partial\Omega,
    \end{cases}
\end{equation}
with differential operator
\begin{equation}
    \mathcal{P}[u] \coloneqq  \tfrac{1}{2}\mathrm{Tr}(\sigma\sigma^\top \mathrm{Hess}_u) + \mu \cdot \nabla u.
\end{equation}
In the above, $\Omega\subset\R^d$ is an open, bounded, connected, and sufficiently regular domain, see, e.g.,~\citet{baldi2017stochastic,karatzas2014brownian,schilling2014brownian} for suitable regularity assumptions. Note that the formulation in~\eqref{eq:pde} includes the Poisson equation for $\mu=0$ and $\sigma = \sqrt{2} \mathrm{I}$, i.e.,
\begin{equation}
\label{eq:pde_poisson}
    \begin{cases}
    \Delta u = f, \quad &\text{on} \quad \Omega, \\
    u = g, &\text{on} \quad \partial\Omega.
    \end{cases}
\end{equation}
In the following, we will summarize existing neural PDE solvers for these PDEs, see~\Cref{tab:compare} for an overview. On a high level, they propose different variational formulations $\min_{v\in V} \, \mathcal{L}[v]$
with the property that the minimizer over a suitable function space $V\subset C(\Omega)$ is a solution $u$ to the PDE in~\eqref{eq:pde}. The space $V$ is then typically approximated by a set of neural networks with a given architecture, such that the minimization problem can be tackled using variants of stochastic gradient descent.

\subsection{Strong and weak formulations of elliptic PDEs}
Let us start with methods based on strong or weak formulations of the PDE in~\eqref{eq:pde}.  

\paragraph{Physics-informed neural networks (PINNs):} 

In its basic form, the loss of PINNs~\cite{raissi2019physics} or \emph{Deep Galerkin} methods~\cite{sirignano2018dgm}, is given by
\begin{equation}
\label{eq:pinn}
    \mathcal{L}_{\mathrm{PINN}}[v] \coloneqq \E\left[(\mathcal{P}[v](\xi) - f(\xi))^2\right] + \beta \mathcal{L}_{\mathrm{\mathrm{bnd}}}[v],
\end{equation}
where 
\begin{equation}
\label{eq:bound_loss}
    \mathcal{L}_{\mathrm{\mathrm{bnd}}}[v] \coloneqq \E_{
    }\left[(v(\zeta) - g(\zeta))^2\right].
\end{equation}
In the above, $\beta \in (0,\infty)$ is a penalty parameter, and $\xi$ and $\zeta$ are suitable random variables distributed on $\Omega$ and $\partial \Omega$, respectively. While improved sampling methods have been investigated, see, e.g.,~\citet{tang2023pinns,chen2023adaptive}, the default choice is to pick uniform distributions. The expectations are then approximated with standard MC or quasi-MC methods based on a suitable set of samples. 

By minimizing the point-wise residual of the PDE, PINNs have gained popularity as a universal and simple method. However, PINNs are sensitive to hyperparameter choices, such as $\beta$, and suffer from training instabilities or high variance~\cite{wang2021understanding, krishnapriyan2021characterizing,nusken2021solving}. Moreover, the objective in~\eqref{eq:pinn} requires the evaluation of the derivatives appearing in $\mathcal{P}[v]$. While this can be done exactly using automatic differentiation, it leads to high computational costs, see~\Cref{fig:mem}.

\paragraph{Deep Ritz method:}

For the Poisson equation in~\eqref{eq:pde_poisson}, one can avoid this cost by leveraging weak variational formulations, 
see, e.g.,~\citet{evans2010partial}. Rather than directly optimizing the regression loss in~\eqref{eq:pinn}, the \emph{Deep Ritz method}~\cite{weinan2017deep} proposes to minimize the objective
\begin{equation}
\label{eq:deepritz}
    \mathcal{L}_{\mathrm{Ritz}}[v] \coloneqq \E\left[\frac{\|\nabla v(\xi)\|^2}{2} - f(\xi)v(\xi)\right] + \beta \mathcal{L}_{\mathrm{\mathrm{bnd}}}[v].
\end{equation}
Under suitable assumptions, the minimizer again corresponds to the solution to the PDE in~\eqref{eq:pde_poisson}. The objective only requires computing the gradient $\nabla v$ instead of the Laplacian $\Delta v$. Using backward mode automatic differentiation, this reduces the number of backward passes from $d+1$ to one, see also the reduced cost in~\Cref{fig:mem}. Moreover, we note that the loss in~\eqref{eq:deepritz} allows for weak solutions that are not twice differentiable. We refer to~\citet{chen2020comparison}, for an extensive comparison of the Deep Ritz method to PINNs for elliptic PDEs with different boundary conditions. 

Finally, we mention that both methods suffer from the fact that the interior losses only consider local, pointwise information at samples $x \in \Omega$. At the beginning of training, the interior loss might thus not be meaningful. Specifically, the boundary condition $g$ first needs to be learned via the boundary loss $\mathcal{L}_{\mathrm{bnd}}$, and then propagate from the boundary $\partial \Omega$ to interior points $x$ via the local interior loss. There exist some heuristics to mitigate this issue by, e.g., progressively learning the solution, see~\citet{penwarden2023unified} for an overview. The next section describes more principled ways of including boundary information in the loss and directly informing the interior points of the boundary condition. 

\subsection{Stochastic formulations of elliptic PDEs}

From weak solutions, we will now proceed to stochastic representations of elliptic PDEs in~\eqref{eq:pde}. To this end, consider the\footnote{We assume that there exists a unique solution, see, e.g.,~\citet{le2016brownian} for corresponding conditions.} solution $X^\xi$ to the stochastic differential equation
\begin{equation}
\label{eq:sde}
    \mathrm{d} X^\xi_t = \mu(X^\xi_t)\, \mathrm{d}t + \sigma(X^\xi_t) \, \mathrm{d} W_t, \quad X^\xi_0 = \xi,
\end{equation}
where $W$ is a standard $d$-dimensional Brownian motion.
Moreover, we define the stopping time $\tau$ as the first exit time of the stochastic process $X^\xi$ from the domain $\Omega$, i.e., 
\begin{equation}
\label{eq:stopping_time}
    \tau = \tau(\Omega, \xi) \coloneqq \inf\{t \in [0, \infty)\colon X_t^\xi \notin \Omega\}.
\end{equation} 
An application of Itô's lemma to the process $u(X^\xi_{t \wedge \tau})$ shows that we almost surely have that
\begin{equation}
    u(X^\xi_{\tau}) = u(X^\xi_0) + \int_{0}^{\tau} \mathcal{P}[u](X^\xi_t)\,  \mathrm{d}t + S^u_\tau,
\end{equation}
where $S_\tau^u$ is the stochastic integral
\begin{equation}
    S_\tau^u \coloneqq \int_{0}^{\tau} (\sigma^\top \nabla u)(X^\xi_t) \cdot  \mathrm{d}W_t.
\end{equation}
Using the fact that $X^\xi_0=\xi$ and assuming that $u$ solves the elliptic PDE in~\eqref{eq:pde},
we arrive at the formula
\begin{equation}
\label{eq:ito}
    g(X^\xi_{\tau}) = u(\xi) + F_\tau^\xi + S_\tau^u,
\end{equation}
where we used the abbreviation
\begin{equation}
   F_\tau^\xi \coloneqq  \int_{0}^{\tau} f(X^\xi_t)\,  \mathrm{d}t.
\end{equation}
Since the stochastic integral $S^\tau_u$ has zero expectation, see, e.g.,~\citet[Theorem 10.2]{baldi2017stochastic}, we can rewrite~\eqref{eq:ito} as a stochastic representation, i.e.,
\begin{equation}
\label{eq:stoch_repr}
    u(x) = \E\left[g(X^\xi_{\tau}) - F_\tau^\xi \big| \xi = x \right],
\end{equation}
which goes back to Kakutani's Theorem~\cite{kakutani1944143} and is a special case of the Feynman-Kac formula.

While the representation in~\eqref{eq:stoch_repr} leads to MC methods for the pointwise approximation of $u$ at a given point $x\in\Omega$, it also allows us to derive variational formulation for learning $u$ on the whole domain $\Omega$. Based on the above results, we can derive the following three losses.

\paragraph{Feynman-Kac loss:}
The \emph{Feynman-Kac} loss is given by
\begin{equation}
\label{eq:fk}    \mathcal{L}_{\mathrm{FK}}[v] \coloneqq \E \Big[ \big( v(\xi) - g(X^\xi_{\tau}) +F_\tau^\xi \big)^2 \Big]
\end{equation}
and follows from the fact that the solution to a quadratic regression problem as in~\eqref{eq:fk} is given by the conditional expectation in~\eqref{eq:stoch_repr}. Notably, this variational formulation does neither require a derivative of the function $v$ nor an extra boundary loss $\mathcal{L}_{\mathrm{bnd}}$.

\paragraph{BSDE loss:}

Since the formula in~\eqref{eq:ito} holds if and only if $u$ solves the PDE in~\eqref{eq:pde}, we can derive the \emph{BSDE} loss 
\begin{equation}
\label{eq:bsde}
    \mathcal{L}_{\mathrm{BSDE}}[v] \coloneqq \E \Big[ \big(v(\xi) - g(X^\xi_{\tau}) +F_\tau^\xi + S_\tau^v \big)^2 \Big].
\end{equation}
Compared to the Feynman-Kac loss in~\eqref{eq:fk}, the BSDE loss requires computing the gradient of $v$ at every time-discretization of the SDE $X^\xi$ in order to compute $S_\tau^v$. However, due to~\eqref{eq:ito}, $S_\tau^v$ acts as a control variates and causes the variance of the MC estimator of~\eqref{eq:bsde} to vanish at the optimum, see~\citet{richter2022robust} for details. 

For the previous two losses, boundary information is directly propagated along the trajectory of the SDE $X^\xi$ to the interior. However, simulating a batch of realizations of the SDE until they reach the boundary $\partial \Omega$, i.e., until the stopping time $\tau$, can incur prohibitively high costs.

\paragraph{Diffusion loss:} The \emph{diffusion loss}~\cite{nusken2021interpolating} circumvents long simulation times by stopping the SDE at $s=\tau \wedge T$, i.e., at the minimum of a prescribed time $T \in (0,\infty)$ and the stopping time $\tau$. Since the trajectories might not reach the boundary, the loss is supplemented with a boundary loss. This yields the variational formulation
\begin{equation}
    \mathcal{L}_{\mathrm{Diff}}[v] \coloneqq \E \big[ \big(v(\xi) - v(X^\xi_{s}) +F^\xi_s + S_{s}^v \big)^2 \big] + \beta \mathcal{L}_{\mathrm{bnd}}[v].
\end{equation}
Note that this can be viewed as an interpolation between the BSDE loss (for $s \to \infty$) and the PINN loss (for $s\to 0$ and rescaling by $s^{-2}$). In the same way, it also balances the advantages and disadvantages of both losses, see also~\Cref{tab:compare}. 

\section{Neural Walk-on-Spheres (NWoS) Method}
\label{sec:nwos}
In this section, we will present a more efficient way of simulating the SDE trajectories for the case of Poisson-type PDEs as in~\eqref{eq:pde_poisson}. Our loss is based on the FK loss in~\eqref{eq:fk}, which does not require the computation of any spatial derivatives of the neural network $v$. However, we reduce the number of steps for simulating the process $X^\xi$ while still reaching the boundary (different from the diffusion loss).

\subsection{Recursion of elliptic PDEs on sub-domains}
\label{sec:recursion}

First, we outline how to cast the solution of the PDE in~\eqref{eq:pde} into nested subproblems of solving elliptic PDEs on subdomains. Specifically, let $\Omega_0\subset \Omega$ be an open sub-domain containing\footnote{Since $\xi$ is a random variable, the sub-domain $\Omega_0$ is random, and the statement is to be understood for each realization.} $\xi_0\coloneqq \xi$ and let $\tau_{0} \coloneqq \tau(\Omega_0, \xi)$ be the corresponding stopping time, defined as in~\eqref{eq:stopping_time}.
Analogously to~\eqref{eq:stoch_repr}, we obtain that
\begin{equation}
\label{eq:stoch_repr_recursive}
    u(\xi) = \E\left[u(X^{\xi}_{\tau_{0}})- F^\xi_{\tau_0} \big| \xi \right].
\end{equation}
Note that this is a recursive definition since the solution $u$ to the PDE in~\eqref{eq:pde} appears again in the expectation. To resolve the recurrence, we define the random variable
     $\xi_1 \sim X^{\xi}_{\tau_0}$
and choose another open sub-domain $\Omega_1 \subset \Omega$ containing $\xi_1$. Considering the stopping time $\tau_{1} \coloneqq \tau(\Omega_1, \xi_1)$, we can calculate the value of $u$ appearing in the inner expectation
\begin{align}
   u(X^{\xi}_{\tau_{0}})\sim u(\xi_1) = \E\left[u(X^{\xi_1}_{\tau_{1}})- F_{\tau_1}^{\xi_1}\big| \xi_{1} \right] 
\end{align}
We can now iterate this process for $k\in\N$ and combine the result with~\eqref{eq:stoch_repr_recursive} to obtain
\begin{equation}
\label{eq:stoch_repr_nested}
    u(\xi) = \E\Bigg[g(X^{\xi}_{\tau})- \sum_{k \ge 0 } F^{\xi_k}_{\tau_{k}} \Bigg| \xi \Bigg].
\end{equation}
In the above, we used the strong Markov property of the SDE solution and the tower property of the conditional expectation, see also~\citet{hermann2020deep}.
This nested stochastic representation can be compared to the one in~\eqref{eq:stoch_repr}.
The next section shows how this provides a practical algorithm that terminates in finitely many steps. 

\subsection{Walk-on-Spheres}
\label{sec:wos}
We tackle the problem of solving the Poisson equation in~\eqref{eq:pde_poisson}, i.e., $\mu=0$ and $\sigma = \sqrt{2} \mathrm{I}$.
Then, the SDE in~\eqref{eq:sde} is just a scaled Brownian motion starting at $\xi$. Picking $\Omega_k \coloneqq B_{r_k}(\xi_k)$ to be a ball of radius $r_k\in (0,\infty)$ around $\xi_k$ in the $k$-th step, the isotropy of Brownian motion ensures that
\begin{equation}
     \xi_{k+1} \sim X^{\xi_{k}}_{\tau_k} \sim \mathcal{U}(\partial B_{r_k}(\xi_k)).
\end{equation}
In other words, we can just sample $\xi_{k+1}$ uniformly from a sphere of radius $r_k$ around the previous value $\xi_k$. To terminate after finitely many steps, we pick the maximal radius in each step, i.e., 
\begin{equation}
r_k \coloneqq \operatorname{dist}(\xi_k,\partial \Omega),   
\end{equation}
and stop at step $\kappa$ when reaching an $\varepsilon$-shell, i.e., when $r_\kappa < \varepsilon$ for a prescribed $\varepsilon\in(0,\infty)$. This allows us to \enquote{walk} from sphere to sphere until (approximately) reaching the boundary, such that we can estimate the first term in~\eqref{eq:stoch_repr_nested}.
Specifically, the value $u(X^\xi_\tau)$ in~\eqref{eq:stoch_repr_nested} is approximated by the boundary value $g(\bar{\xi}_\kappa)$, where 
\begin{equation}
    \bar{\xi}_\kappa \coloneqq \argmin_{x \in \partial \Omega} \|x - \xi_\kappa\|
\end{equation}
is the projection to the boundary.  

We note that the bias from introducing the stopping tolerance $\varepsilon$ can be estimated as $\mathcal{O}(\varepsilon)$~\cite{mascagni2003}. Moreover, for well-behaved, e.g., convex, domains $\Omega$, the average number of steps $\kappa$ behaves like $\mathcal{O}(\log (\varepsilon^{-1}))$ \cite{motoo1959some,binder2012rate}. 
This shows that $\varepsilon$ can be chosen sufficiently small without incurring too much additional computational cost. We note that this leads to much faster convergence than time-discretizations of the Brownian motion. In order to have a comparable bias, we would need to take steps of size $\mathcal{O}(\varepsilon)$, requiring $\Omega(\varepsilon^{-2})$ steps to converge. 

\subsection{Source term}
\label{sec:source}
To compute the second term in~\eqref{eq:stoch_repr_nested}, we need to accumulate values of the form
\begin{equation}
\label{eq:poisson_ball}
    v(z) \coloneqq \E\left[   - F^z_{\tau(B,z)} \right]
\end{equation}
with a given ball $B=B_r(z)$.
By~\eqref{eq:stoch_repr}, we observe that $v$ is the solution of a Poisson equation on the ball $B$ with zero Dirichlet boundary condition evaluated at $z\in\Omega$. We can thus use classical results by~\citet{boggio1905sulle}, see also~\citet{gazzola2010polyharmonic}, to write the solution in terms of Green's functions. Specifically, we have that
\begin{align}
\label{eq:green_repr}
  v(z) &=  -|B_r(z)|\E[f(\gamma)G_r(\gamma,z)],
\end{align}
where $\gamma \sim \mathcal{U}(B_{r}(z))$ and
\begin{equation}
  G_r(y,z)  \coloneqq \begin{cases}
     \frac{1}{2\pi} \log  \frac{r}{\|y-z\|}, &d=2,\\[0.4em]
      \frac{\Gamma(d/2-1)}{4\pi^{d/2}} \left(\|y-z\|^{2-d}-r^{2-d}\right),&d> 2,
  \end{cases}
\end{equation}
see~\Cref{app:green} for further details. In practice, we can now approximate the expectation in~\eqref{eq:green_repr} using an MC estimate. 

\subsection{Learning Problem}
\label{sec:learn}

\begin{figure*}[t]
    \centering
    \includegraphics[width=0.88\linewidth]{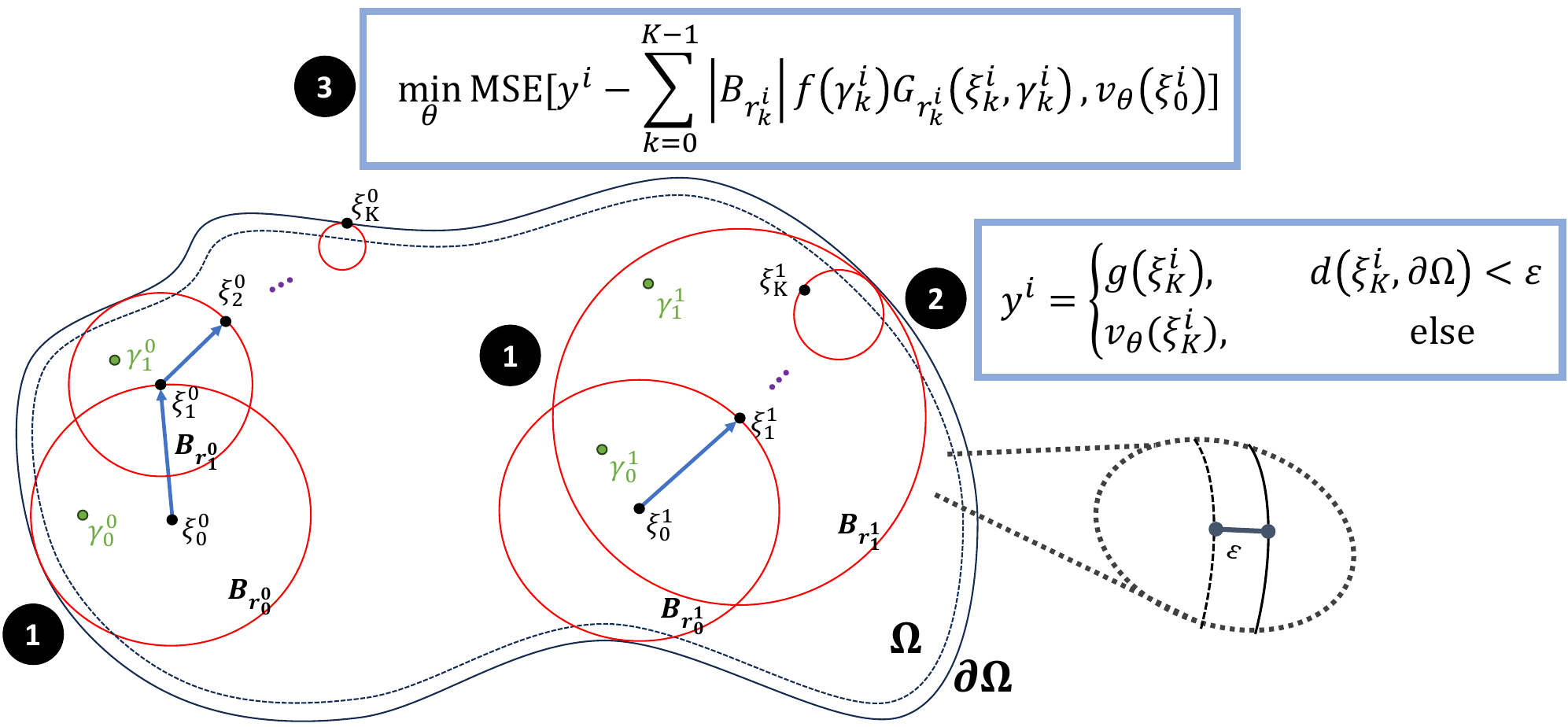}
    \caption{\textbf{Neural Walk-on-Spheres (NWoS):}  Our algorithm for learning the solution to Poisson equations $\Delta u = f$ on $\Omega\subset \R^d$ and $u|_{\partial \Omega} = g$. \protect\circled{1} In each gradient descent step, we sample a batch of random points $(\xi_0^i)_{i=1}^m$ in the domain $\Omega$ and simulate Brownian motions by iteratively sampling $\xi_k^i$ from spheres $B_{r^i_k}$ inscribed in the domain. To account for the source term $f$, we sample $\gamma_k^i \sim \mathcal{U}(B_{r^i_k})$ to compute an MC approximation $|B_{r^i_k}|f(\gamma_k^i)G_{r_k^i}(\xi_k^i,\gamma_k^i)$ to the solution of the Poisson equation on the sphere $B_{r^i_k}$ using the Green's function $G_{r_k^i}$ in \Cref{sec:source}. \protect\circled{2} We stop after a fixed number of maximum steps $K$ and either evaluate our neural network $v_\theta$ or the boundary condition $g$ if we reach an $\varepsilon$-shell of $\partial \Omega$. \protect\circled{3} If $v_\theta$ satisfies the PDE, the mean-value property implies that $v_\theta(\xi_0^i)$ is approximated by the expected value of $y^i$ minus the accumulated source term contributions. We thus minimize the corresponding mean squared error over the parameters $\theta$ using gradient descent.}
    \label{fig:pipeline}
    \vspace{-0.5em}
\end{figure*}

Based on the previous derivations, we can establish a variational formulation, where the minimizer is guaranteed to approximate the Poisson equation in~\eqref{eq:pde_poisson} on the whole domain $\Omega$. Specifically, we define
\begin{equation}
\label{eq:wos_loss}
    \mathcal{L}_{\mathrm{NWoS}}[v] \coloneqq \E \Big[ \big( v(\xi) - \operatorname{WoS}(\xi) \big)^2 \Big],
\end{equation}
where the single-trajectory WoS method $\operatorname{WoS}(\xi)$ with random initial point $\xi$ is given by 
\begin{equation}
    \operatorname{WoS}(\xi) \coloneqq g(\bar{\xi}_\kappa) - \sum_{k=0}^{\kappa-1} |B_{r_k}(\xi_k)|f(\gamma_k)G_{r_k}(\gamma_k,\xi_k).
\end{equation}
In the above, $\gamma_k \sim \mathcal{U}(B_{{r_k}}(\xi_k))$, and the random variables $\kappa$, $\xi_k$, $\bar{\xi}_\kappa$, and $r_k$ are defined as in \Cref{sec:wos}. 
From the stochastic formulation of the solution in~\eqref{eq:stoch_repr_nested} and Proposition 3.5 in~\citet{hermann2020deep}, it follows that the minimizer of~\eqref{eq:wos_loss}, i.e., $x\mapsto \E \big[  \operatorname{WoS}(\xi)\big|\xi = x\big]$, 
approximates the solution $u$ in~\eqref{eq:pde_poisson} in the uniform norm up to error $\mathcal{O}(\varepsilon)$, where $\varepsilon$ is the stopping tolerance, see \Cref{sec:wos}. We also remark that, in theory, the loss requires only a single WoS trajectory per sample of $\xi$ since the minimizer of the regression problem in~\eqref{eq:wos_loss} averages out the noise. 

Having established a learning problem, we can analyze both approximation and generalization errors. For the former,~\citet{hermann2020deep} and~\citet{beznea2022monte} bounded the size of neural networks $v_\theta$ to approximate the solution $u$ up to a given accuracy. In particular, the number of required parameters $\theta$ only scales polynomially in the dimension $d$ and the reciprocal accuracy, as long as the functions $f$, $g$, and $\operatorname{dist}(\cdot, \partial \Omega)$ can be efficiently approximated by neural networks. 

One can then leverage results by~\citet{berner2020analysis} to show that also the generalization error does not underlie the curse of dimensionality when minimizing the empirical risk, i.e., an MC approximation of~\eqref{eq:wos_loss}, over a suitable set of neural networks $v_\theta$. Specifically, the number of required samples of $\xi$ to guarantee that the empirical minimizer approximates the solution $u$ up to a given accuracy also scales only polynomially with dimension and accuracy.

\subsection{Implementation}
\label{sec:implementation}

In this section, we discuss implementations for the loss $\mathcal{L}_{\mathrm{NWoS}}$ in~\eqref{eq:wos_loss} described in the previous section. We summarize our algorithm in \Cref{fig:pipeline} and provide pseudocode for the vanilla version in~\Cref{alg:vanilla_NWoS}. In the following, we present strategies to trade-off accuracy and computational cost and to reduce the variance of MC estimators. We provide pseudocode for NWoS with these improvements in \Cref{alg:NWoS} and \Cref{alg:wos_single} in the appendix.

\begin{table*}[t]
\vskip -6pt
\centering
\caption{Relative $L^2$-error (and standard deviations over $5$ independent runs) of our considered methods, estimated using MC integration on $10^6$ uniformly distributed (unseen) points in $\Omega$.}
\vspace{0.5em}
\begin{tabular}{lrrrr}
\toprule

Method & \multicolumn{4}{c}{Problem} \\ 
 &        Laplace ($10d$)  & Committor ($10d$) & Poisson Rect. ($10d$) & Poisson ($50d$) \\ \midrule
PINN & $7.42e^{-4}\pm1.84e^{-4}$ & $4.10^{-3}\pm1.11e^{-3}$ & $1.35e^{-2}\pm1.57e^{-3}$ & $7.70e^{-3}\pm2.25e^{-3}$ \\ 
Deep Ritz & $8.43e^{-4}\pm6.29e^{-5}$ & $6.15e^{-3}\pm5.30e^{-4}$& $1.06e^{-2}\pm6.20e^{-4}$ & $1.05e^{-3}\pm1.70e^{-4}$ \\ 
Diffusion loss & $1.57e^{-4}\pm7.74e^{-6}$ &$4.48e^{-2}\pm6.93e^{-3}$& $9.69e^{-2}\pm1.03e^{-2}$     & $5.96e^{-4}\pm1.06e^{-5}$  \\ 
Neural Cache & $3.99^{-4}\pm4.08e^{-5}$ &$1.26e^{-3}\pm5.82e^{-5}$& $4.98e^{-2}\pm1.80e^{-2}$     & $1.63e^{-2}\pm1.42e^{-2}$  \\ 
WoS & $1.08e^{-3}\pm1.34e^{-6}$ &$1.99e^{-3}\pm9.79e^{-6}$& $2.32e^{-1}\pm 2.09e^{-1}$ & $4.50e^{-3}\pm7.38e^{-4}$  \\ 
\textbf{NWoS (ours)} & $\mathbf{4.29e^{-5}\pm2.02e^{-6}}$ & $\mathbf{6.56e^{-4}\pm2.42e^{-5}}$  & $\mathbf{2.60e^{-3}\pm9.99e^{-5}}$   & $\mathbf{4.82e^{-4}\pm1.32e^{-5}}$  \\ 
\bottomrule
\end{tabular}
\label{tab:loss}
\vspace{-0.9em}
\end{table*}

\paragraph{WoS with maximum number of steps:}

For sufficiently regular geometries, the probability of a walk taking more than $k$ steps is exponentially decaying in $k$~\cite{binder2012rate}. However, if a single walk in our batch needs significantly more steps, it slows down the overall training. We thus introduce a deterministic maximum number of steps $K\in \mathbb{N}$; see~\citet{beznea2022monte} for a corresponding error analysis. However, we do not want to introduce non-negligible bias by, e.g., just projecting to the closest point on the boundary.

Instead, we want to enforce the mean-value property on subdomains of $\Omega$ based on our recursion in~\Cref{sec:recursion}. We thus propose to use the model $v$ instead of the boundary condition $g$ if the walk does not converge after $K$ steps, i.e., we define\footnote{Since we stop the walk when reaching an $\varepsilon$-shell, the first condition can also be written as $K < \kappa $.}
\begin{equation}
    y^{\xi,v} \coloneqq \begin{cases}
     v(\xi_{K}), & d(\xi_{K}, \partial \Omega ) > \varepsilon,\\
     g(\bar{\xi}_{K}),& \textrm{else.}
      \end{cases}
\end{equation}
We can then replace the second term in~\eqref{eq:wos_loss} by
\begin{equation}
    \operatorname{WoS}(\xi, v) \coloneqq y^{\xi,v} - \sum_{k=0}^{K-1} |B_{r_k}(\xi_k)|f(\gamma_k)G_{r_k}(\gamma_k,\xi_k).
\end{equation}

This helps to reduce the bias when $d(\xi_{K}, \partial \Omega )$ is non-negligible and leads to faster convergence assuming that we obtain increasingly good approximations $v_\theta \approx u$ during training of a neural network $v_\theta$. Our approach bears similarity to the diffusion loss, see~\Cref{sec:methods}; however, we do not need to use a time-discretization of the SDE.

\paragraph{Boundary Loss:}

We find empirically that an additional boundary loss can improve the performance of our method. While theoretically not required, it can especially help for a smaller number $K$ of maximum steps (see the previous paragraph). In general, we thus sample a fraction of the points on the boundary $\partial \Omega$ and optimize
\begin{equation}
    \mathcal{L}_{\mathrm{NWoS}}[v] + \beta \mathcal{L}_{\mathrm{\mathrm{bnd}}}[v],
\end{equation}
where $\mathcal{L}_{\mathrm{\mathrm{bnd}}}$ is defined\footnote{Note that $\mathcal{L}_{\mathrm{\mathrm{bnd}}}$ can be interpreted as a special case of $\mathcal{L}_{\mathrm{NWoS}}$ where the WoS method directly terminates since the initial points are sampled on the boundary.} as in~\eqref{eq:bound_loss}.

\paragraph{Variance-reduction:}
While not necessarily needed for the objective in~\eqref{eq:wos_loss}, we can still average multiple WoS trajectories $N\in\mathbb{N}$ per sample of $\xi$ to reduce the variance. This leads to the estimator
\begin{equation}
\label{eq:estimator}
    \!\!\widehat{\mathcal{L}}_{\mathrm{NWoS}}[v] \coloneqq \frac{1}{m} \left( \sum_{i=1}^m v(\xi^{i}) -\frac{1}{N}\sum_{n=1}^{N} \operatorname{WoS}^{n}(\xi^i) \right)\!,
\end{equation}
where $\xi^i$ are i.i.d.\@ samples of $\xi$ and $\operatorname{WoS}^{n}(\xi^i)$ are i.i.d. samples of $\operatorname{WoS}(\xi^i)$, i.e., $N$ trajectories with the same initial point $\xi^i$, see~\eqref{eq:wos_loss}. Note that we vectorize the WoS simulations across both the initial points and the trajectories, making our NWoS method highly parallelizable and scalable to large batch sizes.

We further introduce control variates to reduce the variance of estimating $\operatorname{WoS}(x)$, where we focus on a fixed $x\in\Omega$ for ease of presentation. Control variates seek to reduce the variance by using an estimator of the form
\begin{equation}
\label{eq:control_variate}
    \E\left[\operatorname{WoS}(x)\right] \approx \E[\delta] + \frac{1}{N}\sum_{n=1}^{N}\operatorname{WoS}^{n}(x) - \delta^{n},
\end{equation}
where $\delta^{n}$ are i.i.d.\@ samples of a random variable $\delta$ with known expectation.

Motivated by~\citet{sawhney2020monte}, we use an approximation of the first-order term of a Taylor series of $u$ in the direction of the first WoS step.
We assume that $\nabla v_\theta$ provides an increasingly accurate approximation of the gradient $\nabla u$ during training and propose to use
\begin{equation}
    \delta^n \coloneqq \nabla v_\theta(x)\cdot(\xi^n_1 - x),
\end{equation}
where $\xi^n_1 $ is the first step of $\operatorname{WoS}^n(x)$. In particular, $\xi^n_1 \sim \mathcal{U}(\partial B_{r_1}(x))$ and thus $\E[\delta]=0$ holds for any function $v_\theta$. 

While we need to compute the gradient $\nabla v_\theta(x)$ for the control variate, we mention that this operation can be detached from the computational graph. In particular, we do not need to compute the derivative of $\nabla v_\theta(x)$ w.r.t.\@ the parameters $\theta$ as is necessary for PINNs, the Deep Ritz method, the diffusion loss, and the BSDE loss. In~\Cref{app:ablation}, we empirically show that the overhead of using the control variates is insignificant. 

\paragraph{Buffer:}
Motivated by~\citet{li2023neural}, we can use a buffer to cache training points 
\begin{equation}
\label{eq:buffer_points}
    \Big(\xi^{(i)}, \frac{1}{N}\sum_{n=1}^N \mathrm{WoS}^n (\xi^{(i)})\Big)_{i=1}^B.
\end{equation}
Since we only update the buffer after a given number of training steps $L\in\mathbb{N}$, this accelerates the training. Note that this is not possible for the other methods since they require evaluation of the current model or its gradients. In every buffer update, we average over additional trajectories, i.e., increase $N$ in~\eqref{eq:buffer_points}, for a fraction of points to improve their accuracy. However, different from~\citet{li2023neural}, we also evict a fraction of points from the buffer and replace them with WoS estimates on newly sampled points $\xi^{(i)}$ in the domain $\Omega$ to balance the diversity and accuracy of the training data in the buffer. 
\begin{remark}
The \emph{Neural Cache} method by \citet{li2023neural} uses a related approach to accelerate WoS methods for applications in computer graphics. However, their method never replaces any point $\xi^{(i)}$ in the buffer, i.e., only updates estimates in the buffer. We observed that the model is thus prone to overfitting on the points in the buffer, especially in high dimensions, preventing it from achieving high accuracies across the domain $\Omega$.
\end{remark}

\begin{algorithm}[t!]
\begin{algorithmic}
\Require neural network $v_\theta$ with initial parameters $\theta$, optimizer method $\operatorname{step}$ for updating the parameters, number of iterations $T$, batch size $m$, source term $f$, boundary term $g$, stopping tolerance $\varepsilon$
\Ensure optimized parameters $\theta$

\For{$k\gets 0,\dots, T$}
\State $ x_{\Omega} \gets \text{sample from } \xi^{\otimes m}$  \Comment{Sample points in $\Omega$}
\State $x \gets x_{\Omega}$ 
\State $ r \gets \operatorname{dist}(x, \partial\Omega)$ \Comment{Compute distances to $\partial\Omega$}
\While{$r > \varepsilon$}
    \State $\gamma \gets \text{sample from } \mathcal{U}(B_{r} (x))$ \Comment{Estimate source}
    \State $ s \gets s - |B_{r}(x)|f(\gamma)G_r(x,\gamma)$  
    \State $ u \gets \text{sample from } \mathcal{U}(\partial B_{r} (x))$
    \State $ x \gets x + u$  \Comment{Walk to next points}
    \State $ r \gets \operatorname{dist}(x, \partial\Omega)$ \Comment{Compute distances to $\partial\Omega$}
    \EndWhile
    \State $x \gets$ project $x$ to $\partial \Omega$ \Comment{Find closest points in $\partial\Omega$}
    \State $y_\Omega \gets s + g(x)$ \Comment{Estimate boundary}

\State $\widehat{\mathcal{L}}_{\mathrm{NWoS}} \gets \operatorname{MSE}(v_\theta(x_{\Omega}), y_{\Omega})$ \Comment{NWoS loss}
\State $\theta \gets \operatorname{step}\big( \gamma, \nabla_\theta \widehat{\mathcal{L}}_{\mathrm{NWoS}} \big)$ \Comment{SGD step}
\EndFor

\end{algorithmic}
\caption{Training of vanilla NWoS method}
\label{alg:vanilla_NWoS}
\end{algorithm}

\section{Experiments}
\label{sec:numerics}

In this section, we compare the performance of NWoS, PINN, DeepRitz, Diffusion loss, and Neural Cache on various problems across dimensions from $10d$ to $50d$. We do not consider the FK and BSDE losses since they incur prohibitively long runtimes for simulating the SDEs with sufficient precision. To compare against the baselines, we consider benchmarks from the works proposing the Deep Ritz and diffusion losses~\citep{jin2017deep,nusken2021interpolating}. For a fair comparison, we set a fixed runtime of $25d + 750$ seconds and GPU memory budget of $2$GiB for training and ran a grid search over a series of hyperparameter configurations for each method. Then, we performed $5$ independent runs for the best configurations w.r.t.\@ the relative $L^2$-error. More details on the hyperparameters and our implementations\footnote{Our PyTorch code can be found at \url{https://github.com/bizoffermark/neural_wos}.} can be found in~\Cref{app:implementation}.

\paragraph{Laplace Equation:}
The first PDE is a Laplace equation on a square domain given by 
\begin{equation}
\textstyle
    f(x) = 0, \quad g(x) = \sum_{i=0}^{d/2} x_{2i}x_{2i+1}, \quad x \in \Omega = (0,1)^d. 
\end{equation}
To test our models, we compare against the analytic solution as $u(x)=\sum_{i=0}^{d/2} x_{2k}x_{2k+1}$. Following~\citet{jin2017deep}, we consider the case $d=10$.

\paragraph{Poisson Equation:}

Next, we consider the Poisson equation presented in~\citet{jin2017deep}, i.e.,
\begin{equation}
\textstyle
    f(x) = 2d, \quad g(x) = \sum_{i=1}^d x_i^2, \quad x \in \Omega = (0,1)^d, 
\end{equation}

with analytic solution $u(x) = \sum_{i=1}^d x_i^2$. We choose  $d=50$ and present results with\footnote{While $d=100$ is considered by~\citet{jin2017deep}, we find that a simple projection outperforms all models in sufficiently high dimensions for this benchmark, see~\Cref{app:additional_eval}.} $d \in \{100,500\}$ in~\Cref{app:additional_eval}. 

\paragraph{Poisson Equation with Rectangular Annulus:} We also consider a Poisson equation on a rectangular annulus $\Omega = (-1,1)^d \setminus [-c, c]^d $
with sinusoidal boundary condition and source term
\begin{equation}
    g(x) = \frac{1}{d} \sum_{i=1}^d \sin(2\pi x_i),  \quad f(x) = -\frac{4\pi^2}{d} \sum_{i=1}^d \sin(2\pi x_i).
\end{equation}
We choose $c=0.25^\frac{1}{d}$ and $d=10$, and note that the analytic solution is given by $u(x) = \frac{1}{d} \sum_{i=1}^d \sin(2\pi x_i)$.
\paragraph{Committor Function: }
The fourth equation deals with \emph{committor functions}
from molecular dynamics. These functions specify likely transition pathways and transition rates between (potentially metastable) regions or conformations of interest~\cite{vanden2006towards,lu2015reactive}. They are typically high-dimensional and known to be challenging to compute. To compare NWoS, we consider the setting in~\citet{nusken2021interpolating}. The task is to estimate the probability of a particle hitting the outer surface of an annulus 
$\Omega = \{x \in \mathbb{R}^d : a<\|x\|<b\}$ with $a, b \in (0,\infty)$,
before the inner surface.

The problem can then be formulated as solving the Laplace equation given by
\begin{equation}
    f(x)=0, \quad g(x) =\mathrm{1}_{\{\|x\| = b\}}, \quad x \in \Omega.
\end{equation}
For this specific $\Omega$, a reference solution can be computed as \begin{equation}
    u(x) = \frac{a^2 - \|x\|^{2-d} a^2}{a^2 - b^{2-d} a^2}.
\end{equation}
We further use the setting by~\citet{nusken2021interpolating} and choose $a=1$, $b=2$, and $d=10$.

\paragraph{PDE-Constrained Optimization: }
Finally, we want to solve the optimization problem
\begin{equation}
\begin{aligned}
&\min_{u \in H_{0}^{1}(\Omega),\, m\in L^2(\Omega)} & \frac{1}{2}\int_{\Omega}(u - u_d)^2dx + \frac{\alpha}{2}\int_\Omega m^2 dx 
\end{aligned}
\end{equation}
constraint to $u$ being a solution to the Poisson equation with $g(x)=0$ and $f(x) = -m(x)$ for $x\in \Omega = (0,1)^2$. The goal of the optimization problem
is to balance the energy of the input control $m$ with the proximity of the state $u$ and the target state $u_d$ while satisfying the PDE constraint. 
Following~\citet{hwang2022solving}, we choose $u_d = \frac{1}{2\pi}\sin(\pi x_1) \sin(\pi x_2)$ as target state.

To tackle this problem and showcase the capabilities of NWoS, we first solve a parametric Poisson equation, where we parametrize the
control as $m_{c} = c_1 \sin(c_2 x_1) \sin(c_3 x_2)$ with
$c \in D\coloneqq [0.5, 1.0] \times [2.5, 3.5]^2$. Similar to~\citet{berner2020numerically}, we can sample random $c\in D$ in every gradient descent step to use NWoS for solving a whole family of Poisson equations. Freezing the trained neural network parameters afterward, we can reduce the PDE-constraint optimization problem to a problem over $c\in D$. In this illustrative example, we can compute the ground-truth parameters as $c^* = \big(\frac{1}{1+4\alpha\pi^4},\pi,\pi\big)$ and choose $\alpha=10^{-3}$.

\begin{figure}[t]
    \centering
    \includegraphics[width=\linewidth]{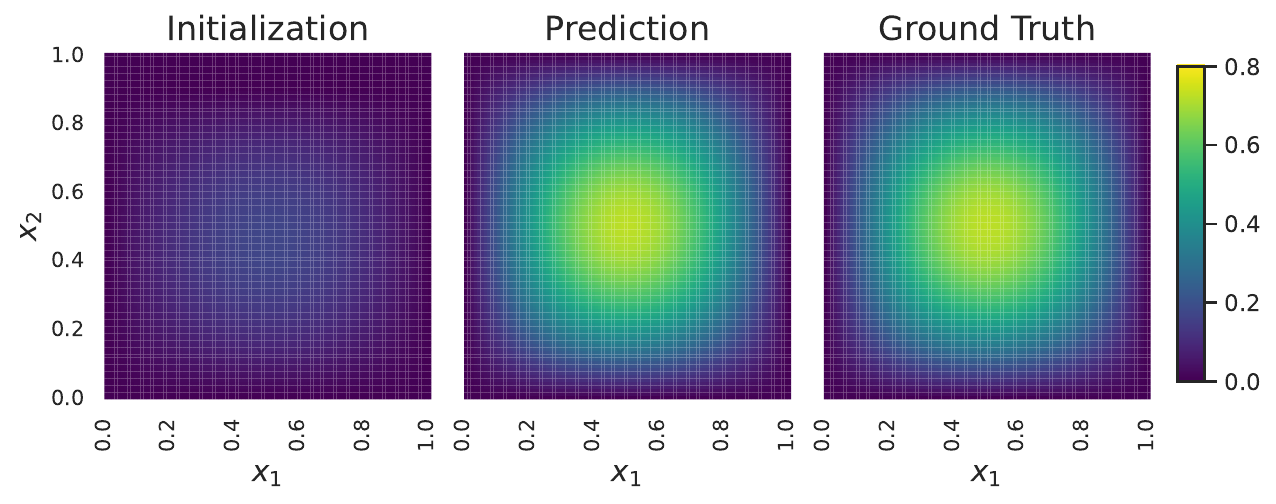}
    \vspace{-1.75em}
    \caption{Qualitative assessment of the solution to the PDE-constrained optimization problem. \textbf{(Left)} Initial function $u_c$ for random parameters $c\in D$. \textbf{(Middle)} Predicted function $u_{\hat{c}}$ for the parameters ${\hat{c}}$ obtained after a few gradient descent steps using the approximation of the solution to the parametric Poisson equation obtained with NWoS. \textbf{(Right)} The groundtruth solution $u_{c^*}$.}
    \label{fig:traj}
    \vspace{-0.5em}
\end{figure}

\subsection{Results}
We present our results in~\Cref{tab:loss}. We first note that we improve the Deep Ritz method and the diffusion loss by almost an order of magnitude compared to the results reported by~\citet{jin2017deep,nusken2021interpolating}. 
Still, our NWoS approach can outperform all other methods on our considered benchmarks. In addition to these results, we highlight that the efficient objective of NWoS also leads to faster convergence, see~\Cref{fig:convergence}. We provide ablation studies in~\Cref{app:ablation} and additional numerical evidence in~\Cref{app:additional_eval}.

The PDE-constrained optimization problem shows that NWoS can be extended to parametric problems, where a whole family of Poisson equations is solved simultaneously. We observe that for this $5$-dimensional problem (two spatial dimensions and three-parameter dimensions), NWoS converges within $20$ minutes to a relative $L^2$-error of $0.79\%$ (averaged over $D\times \Omega$). The trained network can then be used to solve the optimization problem directly (where we use L-BFGS) without requiring an inner loop for the PDE solver. The results show a promising relative $L^2$-error of $1.30$\% for estimating the parameters $c^*$ leading to an accurate prediction of the minimizer, see~\Cref{fig:traj} and~\Cref{app:ablation} for an ablation study.

\section{Conclusion}

We have developed Neural Walk-on-Spheres, a novel way of solving high-dimensional Poisson equations using neural networks. Specifically, we provide a variational formulation with theoretical guarantees that amortizes the cost of the standard Walk-on-Spheres algorithm to learn solutions on the full underlying domain. 
The resulting estimator is more efficient than competing methods (PINNs, the Deep Ritz method, and the diffusion loss) while achieving better performance at lower computational costs and faster convergence. 
We show that NWoS also performs better on a series of challenging, high-dimensional problems and parametric PDEs. This also highlights its potential for applications where such problems are prominent, e.g., in molecular dynamics and PDE-constraint optimization.

\paragraph{Extensions and limitations:}
NWoS is currently only applicable to Poisson equations with Dirichlet boundary conditions. While this PDE appears frequently in applications, we also believe that future work can extend our method. For instance, one can try to leverage adaptations of WoS to spatially varying coefficients~\cite{sawhney2022grid}, drift-diffusion problems~\cite{sabelfeld2017random}, Neumann boundary conditions~\cite{sawhney2023walk,simonov2007random}, fractional Laplacians~\cite{kyprianou2018unbiased}, the screened Poisson or Helmholtz equation~\cite{sawhney2020monte,cheshkova1993walk}, as well as linearized Poisson-Bolzmann equations~\cite{hwang2001efficient,bossy2010probabilistic}. Moreover, one can also take other elementary shapes in each step, e.g., rectangles or stars~\cite{deaconu2006random,sawhney2023walk}, and omit the need for $\varepsilon$-shells for certain geometries using the Green's function first-passage algorithm~\cite{given1997first}.

Finally, while NWoS can tackle parametric PDEs, we need to have a fixed parametrization of the source or boundary functions. It would be promising to extend the ideas to neural operators, which currently only use losses based on PINNs~\cite{goswami2022physics,li2021physics} or diffusion losses for parabolic PDEs~\cite{zhang2023monte}. 

\newpage
\section*{Acknowledgements}
The authors thank Rohan Sawhney for helpful discussions. J.\@ Berner acknowledges support from the Wally Baer and Jeri Weiss Postdoctoral Fellowship. A.\@
Anandkumar is supported in part by Bren endowed chair and by the AI2050 senior fellow program at Schmidt Sciences.

\section*{Impact Statement}
The aim of this work is to advance the field of machine learning and scientific computing. While there are many potential societal consequences of our work, none of them are immediate to require being specifically highlighted here.

\bibliography{refs}

\begin{thebibliography}{63}
\providecommand{\natexlab}[1]{#1}
\providecommand{\url}[1]{\texttt{#1}}
\expandafter\ifx\csname urlstyle\endcsname\relax
  \providecommand{\doi}[1]{doi: #1}\else
  \providecommand{\doi}{doi: \begingroup \urlstyle{rm}\Url}\fi

\bibitem[Azzizadenesheli et~al.(2023)Azzizadenesheli, Kovachki, Li, Liu-Schiaffini, Kossaifi, and Anandkumar]{azzizadenesheli2023neural}
Azzizadenesheli, K., Kovachki, N., Li, Z., Liu-Schiaffini, M., Kossaifi, J., and Anandkumar, A.
\newblock Neural operators for accelerating scientific simulations and design.
\newblock \emph{arXiv preprint arXiv:2309.15325}, 2023.

\bibitem[Bahrami et~al.(2014)Bahrami, Gro{\ss}ardt, Donadi, and Bassi]{bahrami2014schrodinger}
Bahrami, M., Gro{\ss}ardt, A., Donadi, S., and Bassi, A.
\newblock The {Schr{\"o}dinger}--{Newton} equation and its foundations.
\newblock \emph{New Journal of Physics}, 16\penalty0 (11):\penalty0 115007, 2014.

\bibitem[Baldi(2017)]{baldi2017stochastic}
Baldi, P.
\newblock \emph{Stochastic Calculus: An Introduction Through Theory and Exercises}.
\newblock Universitext. Springer International Publishing, 2017.

\bibitem[Beck et~al.(2018)Beck, Becker, Grohs, Jaafari, and Jentzen]{beck2018solving}
Beck, C., Becker, S., Grohs, P., Jaafari, N., and Jentzen, A.
\newblock Solving stochastic differential equations and {K}olmogorov equations by means of deep learning.
\newblock \emph{arXiv preprint arXiv:1806.00421}, 2018.

\bibitem[Beck et~al.(2019)Beck, E, and Jentzen]{beck2019machine}
Beck, C., E, W., and Jentzen, A.
\newblock Machine learning approximation algorithms for high-dimensional fully nonlinear partial differential equations and second-order backward stochastic differential equations.
\newblock \emph{Journal of Nonlinear Science}, 29\penalty0 (4):\penalty0 1563--1619, 2019.

\bibitem[Berner et~al.(2020{\natexlab{a}})Berner, Dablander, and Grohs]{berner2020numerically}
Berner, J., Dablander, M., and Grohs, P.
\newblock Numerically solving parametric families of high-dimensional {Kolmogorov} partial differential equations via deep learning.
\newblock In \emph{Advances in Neural Information Processing Systems}, pp.\  16615--16627, 2020{\natexlab{a}}.

\bibitem[Berner et~al.(2020{\natexlab{b}})Berner, Grohs, and Jentzen]{berner2020analysis}
Berner, J., Grohs, P., and Jentzen, A.
\newblock Analysis of the generalization error: Empirical risk minimization over deep artificial neural networks overcomes the curse of dimensionality in the numerical approximation of {Black--Scholes} partial differential equations.
\newblock \emph{SIAM Journal on Mathematics of Data Science}, 2\penalty0 (3):\penalty0 631--657, 2020{\natexlab{b}}.
\newblock \doi{10.1109/IWOBI.2017.7985525}.

\bibitem[Beznea et~al.(2022)Beznea, Cimpean, Lupascu-Stamate, Popescu, and Zarnescu]{beznea2022monte}
Beznea, L., Cimpean, I., Lupascu-Stamate, O., Popescu, I., and Zarnescu, A.
\newblock From {Monte Carlo} to neural networks approximations of boundary value problems.
\newblock \emph{arXiv preprint arXiv:2209.01432}, 2022.

\bibitem[Binder \& Braverman(2012)Binder and Braverman]{binder2012rate}
Binder, I. and Braverman, M.
\newblock The rate of convergence of the walk on spheres algorithm.
\newblock \emph{Geometric and Functional Analysis}, 22\penalty0 (3):\penalty0 558--587, 2012.

\bibitem[Boggio(1905)]{boggio1905sulle}
Boggio, T.
\newblock Sulle funzioni di green d’ordine m.
\newblock \emph{Rendiconti del Circolo Matematico di Palermo (1884-1940)}, 20:\penalty0 97--135, 1905.

\bibitem[Bossy et~al.(2010)Bossy, Champagnat, Maire, and Talay]{bossy2010probabilistic}
Bossy, M., Champagnat, N., Maire, S., and Talay, D.
\newblock Probabilistic interpretation and random walk on spheres algorithms for the {Poisson-Boltzmann} equation in molecular dynamics.
\newblock \emph{ESAIM: Mathematical Modelling and Numerical Analysis}, 44\penalty0 (5):\penalty0 997--1048, 2010.

\bibitem[Chen et~al.(2020)Chen, Du, and Wu]{chen2020comparison}
Chen, J., Du, R., and Wu, K.
\newblock A comparison study of deep {Galerkin} method and deep {Ritz} method for elliptic problems with different boundary conditions.
\newblock \emph{arXiv preprint arXiv:2005.04554}, 2020.

\bibitem[Chen et~al.(2023)Chen, Cen, and Zou]{chen2023adaptive}
Chen, X., Cen, J., and Zou, Q.
\newblock Adaptive trajectories sampling for solving pdes with deep learning methods.
\newblock \emph{arXiv preprint arXiv:2303.15704}, 2023.

\bibitem[Cheshkova(1993)]{cheshkova1993walk}
Cheshkova, A.
\newblock “walk on spheres” algorithms for solving helmholtz equation.
\newblock \emph{Bulletin of the Novosibirsk Computing Center: Numerical analysis}, \penalty0 (4):\penalty0 7, 1993.

\bibitem[Cuomo et~al.(2022)Cuomo, Di~Cola, Giampaolo, Rozza, Raissi, and Piccialli]{cuomo2022scientific}
Cuomo, S., Di~Cola, V.~S., Giampaolo, F., Rozza, G., Raissi, M., and Piccialli, F.
\newblock Scientific machine learning through physics--informed neural networks: Where we are and what’s next.
\newblock \emph{Journal of Scientific Computing}, 92\penalty0 (3):\penalty0 88, 2022.

\bibitem[De~Ryck \& Mishra(2022)De~Ryck and Mishra]{de2022error}
De~Ryck, T. and Mishra, S.
\newblock Error analysis for physics-informed neural networks {(PINNs)} approximating kolmogorov {PDEs}.
\newblock \emph{Advances in Computational Mathematics}, 48\penalty0 (6):\penalty0 1--40, 2022.

\bibitem[Deaconu \& Lejay(2006)Deaconu and Lejay]{deaconu2006random}
Deaconu, M. and Lejay, A.
\newblock A random walk on rectangles algorithm.
\newblock \emph{Methodology and Computing in Applied Probability}, 8:\penalty0 135--151, 2006.

\bibitem[Duan et~al.(2021)Duan, Jiao, Lai, Lu, and Yang]{duan2021convergence}
Duan, C., Jiao, Y., Lai, Y., Lu, X., and Yang, Z.
\newblock Convergence rate analysis for deep ritz method.
\newblock \emph{arXiv preprint arXiv:2103.13330}, 2021.

\bibitem[E \& Yu(2018)E and Yu]{weinan2018deep}
E, W. and Yu, B.
\newblock The deep ritz method: a deep learning-based numerical algorithm for solving variational problems.
\newblock \emph{Communications in Mathematics and Statistics}, 6\penalty0 (1):\penalty0 1--12, 2018.

\bibitem[E et~al.(2017)E, Han, and Jentzen]{weinan2017deep}
E, W., Han, J., and Jentzen, A.
\newblock Deep learning-based numerical methods for high-dimensional parabolic partial differential equations and backward stochastic differential equations.
\newblock \emph{Communications in Mathematics and Statistics}, 5\penalty0 (4):\penalty0 349--380, 2017.

\bibitem[Evans(2010)]{evans2010partial}
Evans, L.~C.
\newblock \emph{Partial Differential Equations}, volume~19.
\newblock American Mathematical Soc., 2010.

\bibitem[Gazzola et~al.(2010)Gazzola, Grunau, and Sweers]{gazzola2010polyharmonic}
Gazzola, F., Grunau, H.-C., and Sweers, G.
\newblock \emph{Polyharmonic boundary value problems: positivity preserving and nonlinear higher order elliptic equations in bounded domains}.
\newblock Springer Science \& Business Media, 2010.

\bibitem[Given et~al.(1997)Given, Hubbard, and Douglas]{given1997first}
Given, J.~A., Hubbard, J.~B., and Douglas, J.~F.
\newblock A first-passage algorithm for the hydrodynamic friction and diffusion-limited reaction rate of macromolecules.
\newblock \emph{The Journal of chemical physics}, 106\penalty0 (9):\penalty0 3761--3771, 1997.

\bibitem[Goswami et~al.(2022)Goswami, Bora, Yu, and Karniadakis]{goswami2022physics}
Goswami, S., Bora, A., Yu, Y., and Karniadakis, G.~E.
\newblock Physics-informed neural operators.
\newblock \emph{arXiv preprint arXiv:2207.05748}, 2022.

\bibitem[Han et~al.(2017)Han, Jentzen, et~al.]{han2017deep}
Han, J., Jentzen, A., et~al.
\newblock Deep learning-based numerical methods for high-dimensional parabolic partial differential equations and backward stochastic differential equations.
\newblock \emph{Communications in mathematics and statistics}, 5\penalty0 (4):\penalty0 349--380, 2017.

\bibitem[Han et~al.(2018)Han, Jentzen, and E]{han2018solving}
Han, J., Jentzen, A., and E, W.
\newblock Solving high-dimensional partial differential equations using deep learning.
\newblock \emph{Proceedings of the National Academy of Sciences}, 115\penalty0 (34):\penalty0 8505--8510, 2018.

\bibitem[Han et~al.(2020)Han, Nica, and Stinchcombe]{han2020derivative}
Han, J., Nica, M., and Stinchcombe, A.~R.
\newblock A derivative-free method for solving elliptic partial differential equations with deep neural networks.
\newblock \emph{Journal of Computational Physics}, 419:\penalty0 109672, 2020.

\bibitem[Hermann et~al.(2020)Hermann, Sch{\"a}tzle, and No{\'e}]{hermann2020deep}
Hermann, J., Sch{\"a}tzle, Z., and No{\'e}, F.
\newblock Deep-neural-network solution of the electronic {Schr{\"o}dinger} equation.
\newblock \emph{Nature Chemistry}, 12\penalty0 (10):\penalty0 891--897, 2020.

\bibitem[Hwang \& Mascagni(2001)Hwang and Mascagni]{hwang2001efficient}
Hwang, C.-O. and Mascagni, M.
\newblock Efficient modified “walk on spheres” algorithm for the linearized {Poisson--Bolzmann} equation.
\newblock \emph{Applied Physics Letters}, 78\penalty0 (6):\penalty0 787--789, 2001.

\bibitem[Hwang et~al.(2022)Hwang, Lee, Shin, and Hwang]{hwang2022solving}
Hwang, R., Lee, J.~Y., Shin, J.~Y., and Hwang, H.~J.
\newblock Solving {PDE}-constrained control problems using operator learning.
\newblock In \emph{Proceedings of the AAAI Conference on Artificial Intelligence}, volume~36, pp.\  4504--4512, 2022.

\bibitem[Jin et~al.(2017)Jin, McCann, Froustey, and Unser]{jin2017deep}
Jin, K.~H., McCann, M.~T., Froustey, E., and Unser, M.
\newblock Deep convolutional neural network for inverse problems in imaging.
\newblock \emph{IEEE Transactions on Image Processing}, 26\penalty0 (9):\penalty0 4509--4522, 2017.

\bibitem[Juba et~al.(2016)Juba, Keyrouz, Mascagni, and Brady]{juba2016acceleration}
Juba, D., Keyrouz, W., Mascagni, M., and Brady, M.
\newblock Acceleration and parallelization of zeno/walk-on-spheres.
\newblock \emph{Procedia computer science}, 80:\penalty0 269--278, 2016.

\bibitem[Kakutani(1944)]{kakutani1944143}
Kakutani, S.
\newblock Two-dimensional {Brownian} motion and harmonic functions.
\newblock \emph{Proceedings of the Imperial Academy}, 20\penalty0 (10):\penalty0 706--714, 1944.

\bibitem[Karatzas \& Shreve(2014)Karatzas and Shreve]{karatzas2014brownian}
Karatzas, I. and Shreve, S.
\newblock \emph{Brownian motion and stochastic calculus}, volume 113.
\newblock springer, 2014.

\bibitem[Krishnapriyan et~al.(2021)Krishnapriyan, Gholami, Zhe, Kirby, and Mahoney]{krishnapriyan2021characterizing}
Krishnapriyan, A., Gholami, A., Zhe, S., Kirby, R., and Mahoney, M.~W.
\newblock Characterizing possible failure modes in physics-informed neural networks.
\newblock \emph{Advances in Neural Information Processing Systems}, 34, 2021.

\bibitem[Kyprianou et~al.(2018)Kyprianou, Osojnik, and Shardlow]{kyprianou2018unbiased}
Kyprianou, A.~E., Osojnik, A., and Shardlow, T.
\newblock Unbiased ‘walk-on-spheres’ {Monte Carlo} methods for the fractional {Laplacian}.
\newblock \emph{IMA Journal of Numerical Analysis}, 38\penalty0 (3):\penalty0 1550--1578, 2018.

\bibitem[Le~Gall(2016)]{le2016brownian}
Le~Gall, J.-F.
\newblock \emph{Brownian motion, martingales, and stochastic calculus}.
\newblock Springer, 2016.

\bibitem[Li et~al.(2021)Li, Zheng, Kovachki, Jin, Chen, Liu, Azizzadenesheli, and Anandkumar]{li2021physics}
Li, Z., Zheng, H., Kovachki, N., Jin, D., Chen, H., Liu, B., Azizzadenesheli, K., and Anandkumar, A.
\newblock Physics-informed neural operator for learning partial differential equations.
\newblock \emph{arXiv preprint arXiv:2111.03794}, 2021.

\bibitem[Li et~al.(2023)Li, Yang, Deng, De~Sa, Hariharan, and Marschner]{li2023neural}
Li, Z., Yang, G., Deng, X., De~Sa, C., Hariharan, B., and Marschner, S.
\newblock Neural caches for {Monte Carlo} partial differential equation solvers.
\newblock In \emph{SIGGRAPH Asia 2023 Conference Papers}, pp.\  1--10, 2023.

\bibitem[Lu \& Nolen(2015)Lu and Nolen]{lu2015reactive}
Lu, J. and Nolen, J.
\newblock Reactive trajectories and the transition path process.
\newblock \emph{Probability Theory and Related Fields}, 161\penalty0 (1-2):\penalty0 195--244, 2015.

\bibitem[Mascagni \& Hwang(2003)Mascagni and Hwang]{mascagni2003}
Mascagni, M. and Hwang, C.-O.
\newblock $\epsilon$-shell error analysis for “walk on spheres” algorithms.
\newblock \emph{Mathematics and computers in simulation}, 63\penalty0 (2):\penalty0 93--104, 2003.

\bibitem[Miller et~al.(2023)Miller, Sawhney, Crane, and Gkioulekas]{miller2023boundary}
Miller, B., Sawhney, R., Crane, K., and Gkioulekas, I.
\newblock Boundary value caching for walk on spheres.
\newblock \emph{arXiv preprint arXiv:2302.11825}, 2023.

\bibitem[Motoo(1959)]{motoo1959some}
Motoo, M.
\newblock Some evaluations for continuous {Monte Carlo} method by using brownian hitting process.
\newblock \emph{Annals of the Institute of Statistical Mathematics}, 11:\penalty0 49--54, 1959.

\bibitem[Muller(1956)]{muller1956some}
Muller, M.~E.
\newblock Some continuous {Monte Carlo} methods for the dirichlet problem.
\newblock \emph{The Annals of Mathematical Statistics}, pp.\  569--589, 1956.

\bibitem[N{\"u}sken \& Richter(2021{\natexlab{a}})N{\"u}sken and Richter]{nusken2021interpolating}
N{\"u}sken, N. and Richter, L.
\newblock Interpolating between {BSDE}s and {PINN}s: deep learning for elliptic and parabolic boundary value problems.
\newblock \emph{arXiv preprint arXiv:2112.03749}, 2021{\natexlab{a}}.

\bibitem[N{\"u}sken \& Richter(2021{\natexlab{b}})N{\"u}sken and Richter]{nusken2021solving}
N{\"u}sken, N. and Richter, L.
\newblock Solving high-dimensional {Hamilton--Jacobi--Bellman PDEs} using neural networks: perspectives from the theory of controlled diffusions and measures on path space.
\newblock \emph{Partial Differential Equations and Applications}, 2\penalty0 (4):\penalty0 1--48, 2021{\natexlab{b}}.

\bibitem[Penwarden et~al.(2023)Penwarden, Jagtap, Zhe, Karniadakis, and Kirby]{penwarden2023unified}
Penwarden, M., Jagtap, A.~D., Zhe, S., Karniadakis, G.~E., and Kirby, R.~M.
\newblock A unified scalable framework for causal sweeping strategies for physics-informed neural networks {(PINNs)} and their temporal decompositions.
\newblock \emph{arXiv preprint arXiv:2302.14227}, 2023.

\bibitem[Qi et~al.(2022)Qi, Seyb, Bitterli, and Jarosz]{qi2022bidirectional}
Qi, Y., Seyb, D., Bitterli, B., and Jarosz, W.
\newblock A bidirectional formulation for walk on spheres.
\newblock In \emph{Computer Graphics Forum}, volume~41, pp.\  51--62. Wiley Online Library, 2022.

\bibitem[Raissi et~al.(2019)Raissi, Perdikaris, and Karniadakis]{raissi2019physics}
Raissi, M., Perdikaris, P., and Karniadakis, G.~E.
\newblock Physics-informed neural networks: A deep learning framework for solving forward and inverse problems involving nonlinear partial differential equations.
\newblock \emph{Journal of Computational Physics}, 378:\penalty0 686--707, 2019.

\bibitem[Richter \& Berner(2022)Richter and Berner]{richter2022robust}
Richter, L. and Berner, J.
\newblock Robust {SDE}-based variational formulations for solving linear {PDE}s via deep learning.
\newblock In \emph{Proceedings of the 39th International Conference on Machine Learning}, volume 162 of \emph{Proceedings of Machine Learning Research}, pp.\  18649--18666. PMLR, 2022.

\bibitem[Sabelfeld(2017)]{sabelfeld2017random}
Sabelfeld, K.~K.
\newblock Random walk on spheres algorithm for solving transient drift-diffusion-reaction problems.
\newblock \emph{Monte Carlo Methods and Applications}, 23\penalty0 (3):\penalty0 189--212, 2017.

\bibitem[Sawhney \& Crane(2020)Sawhney and Crane]{sawhney2020monte}
Sawhney, R. and Crane, K.
\newblock {Monte Carlo} geometry processing: A grid-free approach to {PDE}-based methods on volumetric domains.
\newblock \emph{ACM Transactions on Graphics}, 39\penalty0 (4), 2020.

\bibitem[Sawhney et~al.(2022)Sawhney, Seyb, Jarosz, and Crane]{sawhney2022grid}
Sawhney, R., Seyb, D., Jarosz, W., and Crane, K.
\newblock Grid-free {Monte Carlo} for {PDE}s with spatially varying coefficients.
\newblock \emph{ACM Transactions on Graphics (TOG)}, 41\penalty0 (4):\penalty0 1--17, 2022.

\bibitem[Sawhney et~al.(2023)Sawhney, Miller, Gkioulekas, and Crane]{sawhney2023walk}
Sawhney, R., Miller, B., Gkioulekas, I., and Crane, K.
\newblock Walk on stars: A grid-free {Monte Carlo} method for {PDE}s with {Neumann} boundary conditions.
\newblock \emph{arXiv preprint arXiv:2302.11815}, 2023.

\bibitem[Scherbela et~al.(2022)Scherbela, Reisenhofer, Gerard, Marquetand, and Grohs]{scherbela2022solving}
Scherbela, M., Reisenhofer, R., Gerard, L., Marquetand, P., and Grohs, P.
\newblock Solving the electronic {Schr{\"o}dinger} equation for multiple nuclear geometries with weight-sharing deep neural networks.
\newblock \emph{Nature Computational Science}, 2\penalty0 (5):\penalty0 331--341, 2022.

\bibitem[Schilling \& Partzsch(2014)Schilling and Partzsch]{schilling2014brownian}
Schilling, R.~L. and Partzsch, L.
\newblock \emph{Brownian motion: an introduction to stochastic processes}.
\newblock Walter de Gruyter GmbH \& Co KG, 2014.

\bibitem[Simonov(2007)]{simonov2007random}
Simonov, N.
\newblock Random walk-on-spheres algorithms for solving mixed and {Neumann} boundary-value problems.
\newblock \emph{Sibirskii Zhurnal Vychislitel'noi Matematiki}, 10\penalty0 (2):\penalty0 209--220, 2007.

\bibitem[Sirignano \& Spiliopoulos(2018)Sirignano and Spiliopoulos]{sirignano2018dgm}
Sirignano, J. and Spiliopoulos, K.
\newblock {DGM}: A deep learning algorithm for solving partial differential equations.
\newblock \emph{Journal of computational physics}, 375:\penalty0 1339--1364, 2018.

\bibitem[Tang et~al.(2023)Tang, Wan, and Yang]{tang2023pinns}
Tang, K., Wan, X., and Yang, C.
\newblock {DAS-PINN}s: A deep adaptive sampling method for solving high-dimensional partial differential equations.
\newblock \emph{Journal of Computational Physics}, 476:\penalty0 111868, 2023.

\bibitem[Vanden-Eijnden et~al.(2006)]{vanden2006towards}
Vanden-Eijnden, E. et~al.
\newblock Towards a theory of transition paths.
\newblock \emph{Journal of statistical physics}, 123\penalty0 (3):\penalty0 503--523, 2006.

\bibitem[Wang et~al.(2021)Wang, Teng, and Perdikaris]{wang2021understanding}
Wang, S., Teng, Y., and Perdikaris, P.
\newblock Understanding and mitigating gradient flow pathologies in physics-informed neural networks.
\newblock \emph{SIAM Journal on Scientific Computing}, 43\penalty0 (5):\penalty0 A3055--A3081, 2021.

\bibitem[Zhang et~al.(2023{\natexlab{a}})Zhang, Meng, Zhu, Wang, Shi, Zhang, Ma, and Liu]{zhang2023monte}
Zhang, R., Meng, Q., Zhu, R., Wang, Y., Shi, W., Zhang, S., Ma, Z.-M., and Liu, T.-Y.
\newblock {Monte Carlo} neural operator for learning pdes via probabilistic representation.
\newblock \emph{arXiv preprint arXiv:2302.05104}, 2023{\natexlab{a}}.

\bibitem[Zhang et~al.(2023{\natexlab{b}})Zhang, Wang, Helwig, Luo, Fu, Xie, Liu, Lin, Xu, Yan, et~al.]{zhang2023artificial}
Zhang, X., Wang, L., Helwig, J., Luo, Y., Fu, C., Xie, Y., Liu, M., Lin, Y., Xu, Z., Yan, K., et~al.
\newblock Artificial intelligence for science in quantum, atomistic, and continuum systems.
\newblock \emph{arXiv preprint arXiv:2307.08423}, 2023{\natexlab{b}}.

\end{thebibliography}
\bibliographystyle{icml2024}

\newpage
\appendix

\clearpage

\section{Green's function for the Ball}
\label{app:green}

For the sake of completeness, this section provides details on the derivation in~\Cref{sec:source}. To compute integrals of the form~\eqref{eq:poisson_ball}, we look at a special case of a Poisson equation on a ball $B=B_r(z)$ with zero Dirichlet boundary condition, i.e.,
\begin{equation}
\label{eq:pde_poisson_ball}
    \begin{cases}
    \Delta v = f, \quad &\text{on} \quad B, \\
    v = 0, &\text{on} \quad \partial B.
    \end{cases}
\end{equation}
Analogously to~\eqref{eq:stoch_repr}, we obtain that 
\begin{equation}
\label{eq:poisson_ball_app}
    v(z) = \E\left[   - \int_{0}^{\tau(B,z)} f(X^z_t)\,  \mathrm{d}t\right],
\end{equation}
where $\tau(B,z)$ is the corresponding stopping time, see~\eqref{eq:stopping_time}. However, since we simplified the domain to a simple ball, we can write the solution in terms of Green's functions. Specifically, we have that
\begin{align}
\label{eq:green_repr_app}
  v(z) &= -\int_{B} f(y) G_r(y,z) \, \mathrm{d}y 
\end{align}
where
\begin{equation}
  G_r(y,z)  \coloneqq \begin{cases}
     \frac{1}{2\pi} \log  \frac{r}{\|y-z\|}, &d=2,\\[0.4em]
      \frac{\Gamma(d/2-1)}{4\pi^{d/2}} \left(\|y-z\|^{2-d}-r^{2-d}\right),&d> 2.
  \end{cases}
\end{equation}
We note that~\eqref{eq:green_repr_app} is equivalent to~\eqref{eq:green_repr}. 

While this is a classical result by~\citet{boggio1905sulle}, see also~\citet{gazzola2010polyharmonic}, we will sketch a proof in the following. We consider the Laplace equation $\Delta \Phi_x = \delta_x$ for given $x\in\R^d$ in the distributional sense. It is well known that the fundamental solution $\Phi_x$ is given by
\begin{equation}
    \Phi_x(y) = \begin{cases}
     \frac{1}{2\pi} \log  \|y-x\|, &d=2,\\
      -\frac{\|y-x\|^{2-d}}{(d-2)\omega_d} ,&d> 2,
  \end{cases}
\end{equation}
where 
\begin{equation}
    \omega_d = |\partial B_1(0)| = \frac{2\pi^{d/2}}{\Gamma(d/2)} = \frac{4\pi^{d/2}}{(d-2)\Gamma(d/2 - 1)}
\end{equation}
is the surface measure of the d-dimensional unit ball $B_1(0)$.
Under suitable conditions, it further holds that the solution to~\eqref{eq:poisson_ball_app} is given by
\begin{equation}
\label{eq:green_corrector}
    v(x) = \int_{B} f(y) \left(  \Phi_x(y) - \phi_x(y) \right)\, \mathrm{d}y
\end{equation}
for every $x \in B$, where the \emph{corrector function} $\phi_x$ satisfies the Laplace equation
\begin{equation}
\label{eq:pde_corrector}
    \begin{cases}
    \Delta \phi_x = 0, \quad &\text{on} \quad B, \\
    \phi_x = \Phi_x, &\text{on} \quad \partial B,
    \end{cases}
\end{equation}
see~\citet[Chapter 2.2]{evans2010partial}.
Based on~\eqref{eq:stoch_repr} and the fact that $\Phi_z$ is constant at the boundary of $B=z + B_r(0)$, we can compute the value of the corrector function $\phi_z$, i.e.,
\begin{align}
    \phi_z(y) = \E[\Phi_z(X^y_{\tau(B,y)})] =\begin{cases}
     \frac{1}{2\pi} \log  r, &d=2,\\[0.4em]
      -\frac{r^{d-2}}{(2-d)\omega_d} ,&d> 2.
  \end{cases}
\end{align}
This shows that the value of the Green's function at the center $z$ of the ball $B$ is given by
\begin{equation}
   \Phi_z(y) - \phi_z(y)  = -G_r(y,z),
\end{equation}
which, together with~\eqref{eq:green_corrector}, establishes the claim.
\subsection{Stable Implementation}
For numerical stability, we directly compute the quantity $\tilde{G}_r(\gamma, z) \coloneqq |B_r(z)|G_r(\gamma, z)$ in practice, as needed in~\eqref{eq:green_repr}. The volume of the hyper-sphere $|B_{r}(z)|$ is given by 
\begin{equation}
    |B_{r}(z)| = \frac{\pi^{\frac{d}{2}}}{\Gamma(\frac{d}{2}+1)} r^d,
\end{equation}
such that we obtain
\begin{equation}
  \tilde{G}_r(\gamma, z) \coloneqq  \coloneqq \begin{cases}
     \frac{r^2}{2} \log  \frac{r}{\|\gamma-z\|}, &d=2,\\[0.4em]
      \frac{r^d}{d(d-2)} \left(\|\gamma-z\|^{2-d}-r^{2-d}\right),&d> 2.
  \end{cases}
\end{equation}

\begin{algorithm}[t!]
\begin{algorithmic}
\Require neural network $v_\theta$ with initial parameters $\theta$, optimizer method $\operatorname{step}$ for updating the parameters, WoS method $\operatorname{WoS}$ in~\Cref{alg:wos_single}, number of iterations $T$, batch sizes $m_d$ and $m_b$ for domain and boundary points, buffer $\mathcal{B}$ of size $B$, boundary function $g$, buffer update interval $L$, boundary penalty parameter $\beta$
\Ensure optimized parameters $\theta$
\State $ x_{\partial\Omega} \gets\text{sample from } \zeta^{\otimes B}$ \Comment{Sample points in $\partial\Omega$}
\State $\mathcal{B} \gets \text{initialize with } (x_{\partial\Omega}, g(x_{\partial\Omega}) )$ \Comment{Initialize buffer}

\For{$k\gets 0,\dots, T$}
\If{$k \!\! \mod L = 0$}
\State $ x_{\Omega} \gets \text{sample from } \xi^{\otimes m_d}$  \Comment{Sample points in $\Omega$}
\State $ x_{\mathcal{B}} \gets \text{sample from } \mathcal{B} $ \Comment{Sample points in $\mathcal{B}$}
\State $x \gets [x_{\Omega}, x_{\mathcal{B}}]$ \Comment{Concatenate points}
\State $ [y_\Omega, y_{\mathcal{B}}] \gets \operatorname{vmap}[\operatorname{WoS}(x, v_\theta)] $ \Comment{WoS}
\State $\mathcal{B} \gets \textrm{update with } (x_{\mathcal{B}}, y_{\mathcal{B}})$ \Comment{Update estimates}
\State $ \mathcal{B} \gets \text{replace with } (x_\Omega, y_\Omega)$ \Comment{Replace points}
\EndIf
\State $ x_{\partial\Omega} \gets \text{sample from } \zeta^{\otimes m_b}$ \Comment{Sample points in $\partial\Omega$}
\State $ (x_{\mathcal{B}},y_{\mathcal{B}}) \gets \text{sample from } \mathcal{B} $ \Comment{Sample points in $\mathcal{B}$}
\State $\widehat{\mathcal{L}}_{\mathrm{NWoS}} \gets \operatorname{MSE}(v_\theta(x_{\mathcal{B}}), y_{\mathcal{B}})$ \Comment{Domain loss}
\State $\widehat{\mathcal{L}}_{\mathrm{bnd}} \gets \operatorname{MSE}(v_\theta(x_{\partial\Omega}), g(x_{\partial\Omega}))$ \Comment{Boundary loss}

\State $ \widehat{\mathcal{L}} =  \widehat{\mathcal{L}}_{\mathrm{NWoS}} + \beta \widehat{\mathcal{L}}_{\mathrm{bnd}}$ 
\State $\theta \gets \operatorname{step}\big( \gamma, \nabla_\theta \widehat{\mathcal{L}} \big)$ \Comment{SGD step}

\EndFor
\end{algorithmic}
\caption{Training of our NWoS method}
\label{alg:NWoS}
\end{algorithm}

\begin{algorithm}[t!]
\begin{algorithmic}
\Require neural network $v_{\theta}$, source term $f$, boundary term $g$, point for evaluation $x$, maximum number of steps $K$, stopping tolerance $\varepsilon$, number of trajectories $N$
\Ensure estimator $\widehat{v}$ of solution $v$ to PDE in~\eqref{eq:pde_poisson} at $x$
\State ${\hat{v} \gets 0}$
\For{$i\gets 1,\dots,N$} \Comment{Batched in implementation}
    \State $s \gets 0$
    \For{$t\gets 1,\dots, K$} 
    \State $ r \gets \operatorname{dist}(x, \partial\Omega)$ \Comment{Compute distance to $\partial\Omega$}
    \If{$r < \varepsilon$}
        \State Break \Comment{Reach boundary}
    \EndIf
    \State $\gamma \gets \text{sample from } \mathcal{U}(B_{r} (x))$ \Comment{Estimate source}
    \State $ s \gets s - |B_{r}(x)|f(\gamma)G_r(x,\gamma)$  
    \State $ u \gets \text{sample from } \mathcal{U}(\partial B_{r} (x))$
    \If{$t = 0 \And{} \mathrm{use\_control\_variate}$}
        \State $s \gets s - \nabla_x v_\theta(x)\cdot u $ \Comment{Control variate}
    \EndIf
    \State $ x \gets x + u$  \Comment{Walk to next point}
    \EndFor
    \If{$r < \varepsilon$}    \Comment{Estimate solution at $x$}
        \State $x \gets$ project $x$ to $\partial \Omega$ \Comment{Find closest point in $\partial\Omega$}
        \State $\widehat{v} \gets s + g(x)$
    \Else
        \State $\widehat{v} \gets s + v_\theta(x)$
    \EndIf
\EndFor
\State $\hat{v} \gets \frac{1}{N} \hat{v}$ \Comment{Compute MC estimate}
\end{algorithmic}
\caption{Walk-on-Spheres ($\operatorname{WoS}$)}
\label{alg:wos_single}
\end{algorithm}

\section{Implementation Details}

\label{app:implementation}
We implemented all methods in PyTorch and provide pseudocode in~\Cref{alg:NWoS,alg:wos_single}. The experiments have been conducted on A100 GPUs. 

For all our training, we use the Adam optimizer and limit the runtime to $25d + 750$ seconds for a fair comparison. In every step, we sample uniformly distributed samples $(\xi,\zeta)$ in the domain $\Omega$ and on the boundary $\partial \Omega$ to approximate the expectations of the loss and boundary terms.  Moreover, we employ an exponentially decaying learning rate, which reduces the initial learning rate by two orders of magnitude throughout training. We choose a feedforward neural network with residual connections, $6$ layers, a width of $256$, and a GELU activation function. We also perform the grid search for the boundary loss penalty term, i.e.,
\begin{equation}
    \beta\in \{0.5, 1, 5, 50, 100, 500, 1000, 5000\}.
\end{equation}
We further include the batch size $m\in \{ 2^{i} \}_{i=7}^{17}$ in our grid search. For a fair comparison, we set a fixed GPU memory budget of $2$GiB for training, leading to different maximal batch-sizes depending on the method; see also \Cref{fig:mem}. Unless otherwise specified, $10\%$ of the batch size is used for boundary points. Moreover, we set $\varepsilon=10^{-4}$ for all methods using an $\varepsilon$-shell. Let us detail the hyperparameter choices specific to each method in the following.

\paragraph{Walk-on-Spheres (WoS):} For WoS~\cite{muller1956some}, we directly approximate the solution at the evaluation points. We batch trajectories to saturate the memory budget and present the best result for different configurations within the given runtime. Specifically, we pick the number of trajectories $N$ in the grid $\{1,10,100,1000,10000,100000\}$ and the maximum number of steps $K$ in $\{0,1,10,100,1000\}$.

\paragraph{Neural Walk-on-Spheres (NWoS):} For NWoS, we try the different extensions in~\Cref{sec:implementation}. Specifically, we fix the buffer size $B$ to $10$ times that of the batch size $m$, and sweep the number of gradient steps between buffer updates $L\in \{10,100,1000\}$. We also include the maximum number of WoS steps $K\in \{0,1,5,10,50,100\}$ and the number of trajectories per update $N\in \{1,10,100,200,300,400,500,1000\}$ in our grid search. If using a boundary loss, we sweep over $\{0.1, 0.2, 0.3, 0.4, 0.5\}$ in the grid search to find the optimal proportion of the batch size for the boundary loss.

\paragraph{Neural Cache:} For the neural cache method~\cite{li2023neural}, we use the best configuration for different buffer sizes, update intervals, and number of trajectories within the given time and memory constraints. Specifically, we try buffer sizes $B\in\{10000,20000,100000, 1000000 \}$, intervals $L\in\{1,10,100,1000,5000,10000\}$ to update the buffer, and $N\in \{1,10,20,30,40,50,100,500,1000\}$ number of trajectories for each update.

\paragraph{Diffusion loss:} For the diffusion loss~\cite{nusken2021interpolating}, we perform a grid search over the time-steps $\Delta t \in \{10^{-3},10^{-4},10^{-5}\}$ of the Euler-Maruyama scheme and the maximum number of steps in $\{1,5,10,50\}$.

\paragraph{PINNs:} For PINNs~\cite{raissi2019physics,sirignano2018dgm}, we use automatic differentiation to compute the Laplacian $\Delta v_\theta$.

\paragraph{Deep Ritz:} For the Deep Ritz method~\cite{weinan2017deep}, we experiment with the original network architecture proposed in their paper. We sweep the number of blocks in $\{4,6,8\}$, the number of layers in $\{2,4\}$, and the hidden dimension in $\{64,128,256\}$. Moreover, we replace the activation function with GELU.

\section{Ablation Studies}
\label{app:ablation}

\begin{table}[t]
    \vskip -6pt

    \caption{Ablation study of the contribution of control variates and the neural network evaluation for trajectories that did not converge after a given maximum number of steps $K$. We report the relative $L^2$-error for the parameter estimation in our PDE-constrained optimization problem.}
    \vspace{0.5em}
    \centering
    \begin{tabular}{lr}
        \toprule
         Method &  Relative $L^2$-error \\ \midrule
        Base NWoS & 2.89\% \\
        + Control Variate & 1.72\%  \\
        + Terminal Eval \quad  & 1.70\% \\
        + Both & \textbf{1.30\%} \\ \bottomrule
    \end{tabular}
    \label{tab:ablation}
\end{table}

\begin{figure}[ht!]
    \centering
    \includegraphics[width=0.78\linewidth]{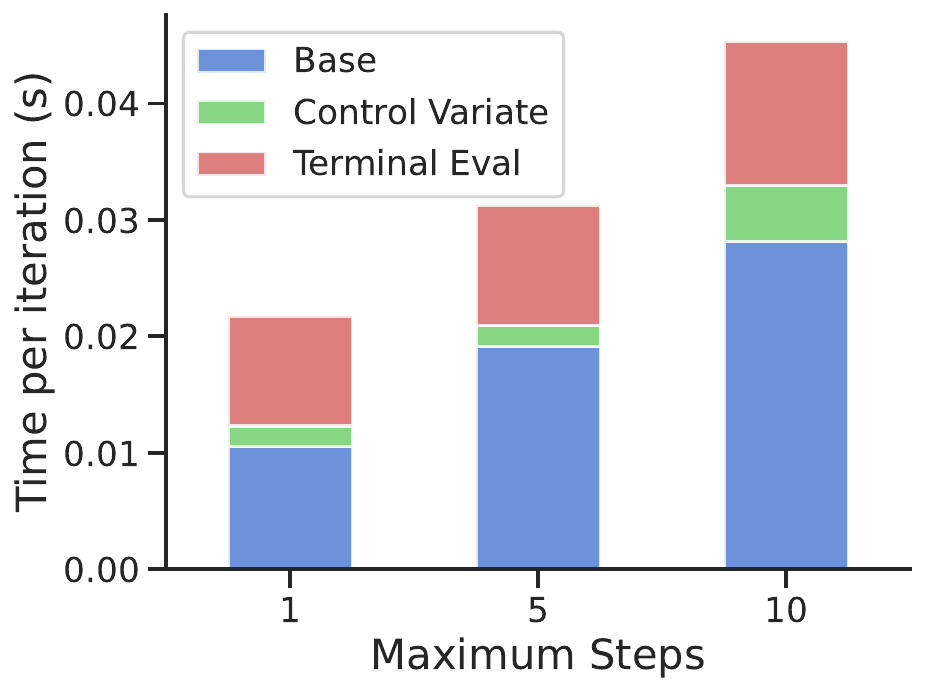}
    \vspace{-0.6em}
    \caption{Decomposition of the training time for one iteration of NWoS in the plain version (\Cref{sec:learn}), as well as using our improvements from~\Cref{sec:implementation}, i.e., the control variates and a neural network evaluation for trajectories that did not converge after a given maximum number of steps $K$.}
    \label{fig:speed_decomposition}
\end{figure}

\begin{figure}[ht!]
    \centering
    \includegraphics[width=0.85\linewidth]{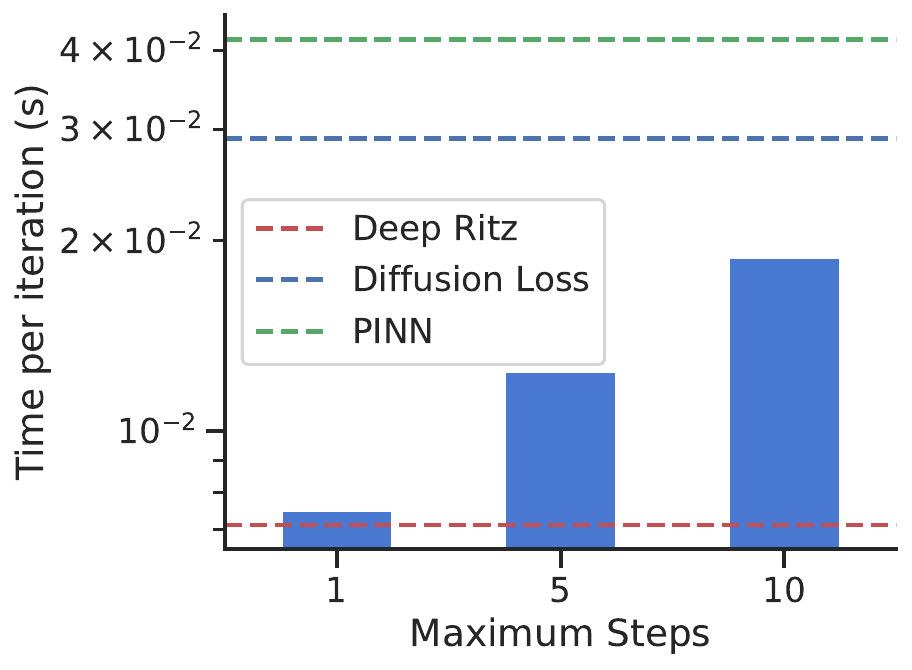}
    \vspace{-0.75em}
    \caption{Training time of our considered methods for one gradient step. For NWoS, we present the comparison for a different maximum number of steps $K$.}
    \vspace{-0.5em}
    \label{fig:speed_poisson}
\end{figure}

In this section, we provide additional ablation studies on the the contribution of our additional improvements in~\Cref{sec:implementation}, namely the control variates as well as the neural network evaluation for trajectories that did not reach the boundary. We also analyze the speed of NWoS and perform comparisons with PINNs, Deep Ritz, and the diffusion loss.

In~\Cref{tab:ablation}, we perform an ablation study on the contribution of our improvements in our PDE-constrained optimization problem. We observe that they can decrease the relative $L^2$-error from $2.89$\% to $1.72$\% and $1.70 $\%, respectively. If both the control variate and the neural network evaluation are used, we obtain the best relative error of $1.30$\%, indicating that they can efficiently decrease bias and variance.  

Figure~\ref{fig:speed_decomposition} decomposes the training time per iteration into the time for the base NWoS algorithm and the time for the additional extensions from~\Cref{sec:implementation}. We assume the batch size to be fixed to $m=512$ and test on the Poisson equation in~\Cref{sec:implementation} in $100d$. We observe that our proposed extensions incur comparably small overheads.
Figure~\ref{fig:speed_poisson} further compares NWoS with DeepRitz, NSDE, and PINN with different maximum number of steps $K$, see~\Cref{sec:implementation}. Considering the logarithmic scaling of the plot, each iteration of NWoS is significantly faster than PINN and NSDE but slower than Deep Ritz for a larger maximum number of steps $K$. Choosing $K$, we can balance high accuracy and fast training.

\section{Further Evaluations}
\label{app:additional_eval}

In this section, we provide further numerical evidence. We report the convergence of the relative $L^2$-error for the other PDEs and evaluate our method on the Poisson equation in $100d$ and $500d$.

\begin{figure}[ht!]
    \centering
    \centering
    \includegraphics[width=0.85\linewidth]{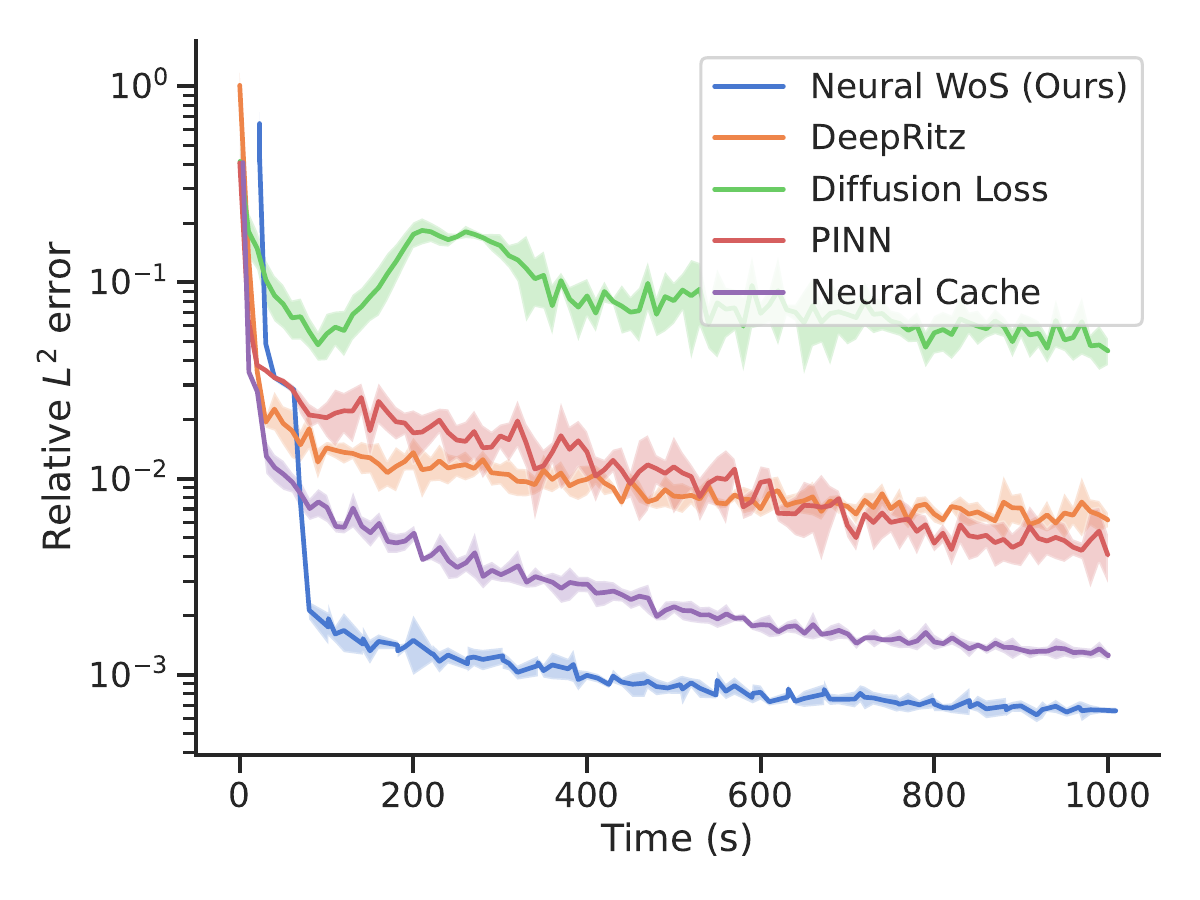}   
    \vspace{-1em}
    \caption{Convergence of the relative $L^2$-error when solving the Committor function in 10$d$ using our considered methods.}
    \label{fig:committor_timed}
    \vspace{-0.5em}
\end{figure}

\begin{figure}[ht!]
    \centering
    \includegraphics[width=0.85\linewidth]{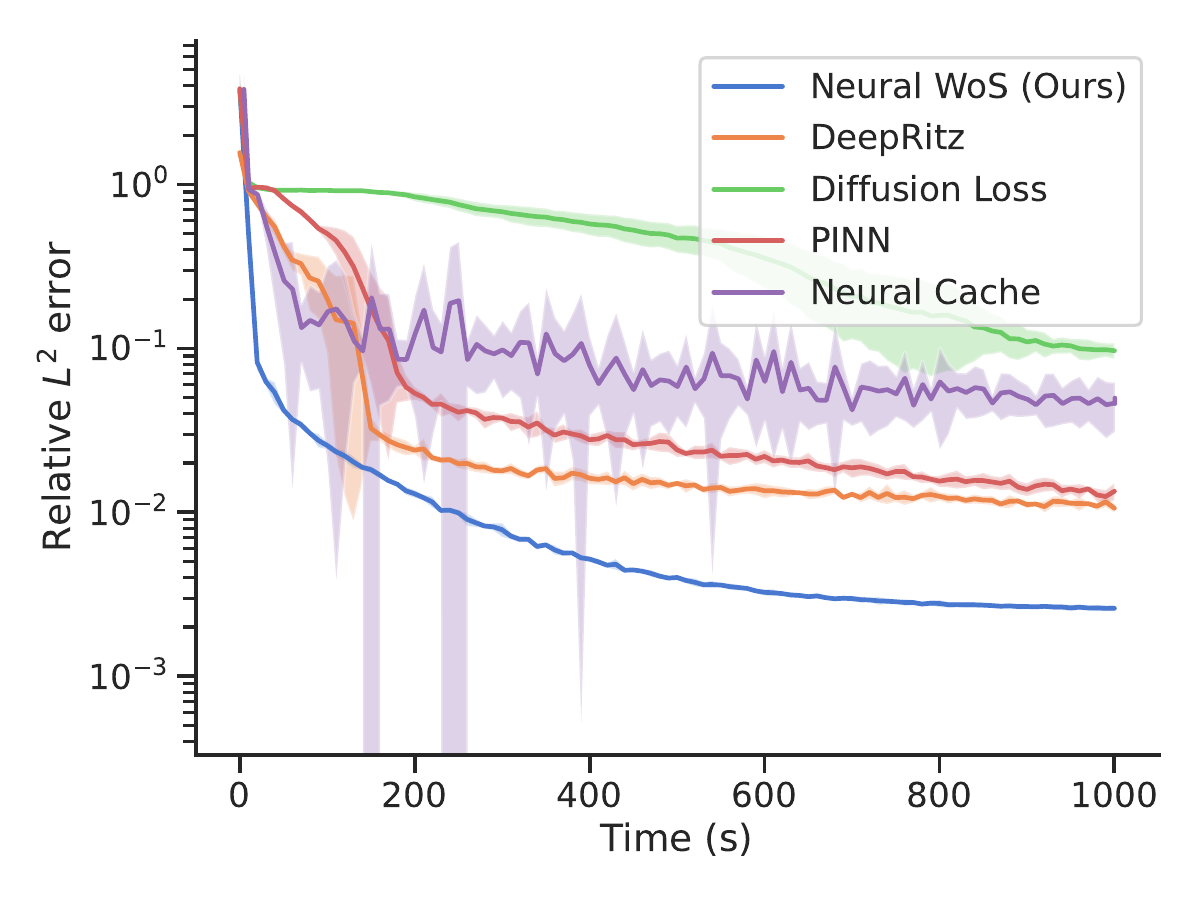}
    \vspace{-1em}
    \caption{Convergence of the relative $L^2$-error when solving the Poisson equation in $10d$ with rectangular torus using our considered methods.}
    \label{fig:poisson_10}
    \vspace{-0.5em}
\end{figure}

\begin{figure}[ht!]
    \centering
    \includegraphics[width=0.85
    \linewidth]{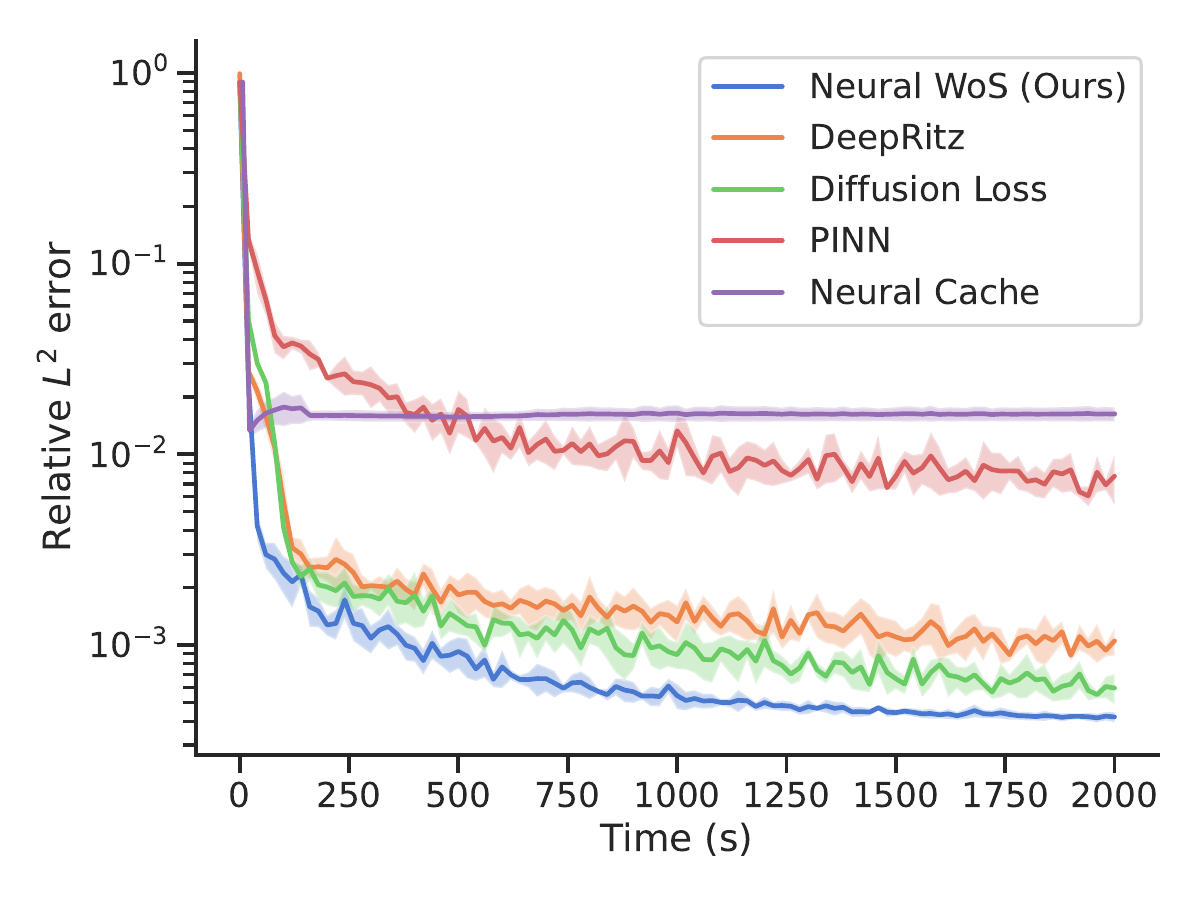}
    \vspace{-1em}
    \caption{Convergence of the relative $L^2$-error when solving the Poisson equation in $50d$ using our considered methods.}
    \label{fig:poisson_50}
    \vspace{-0.5em}
\end{figure}

\Cref{fig:committor_timed,fig:poisson_10,fig:poisson_50} demonstrate that neural WoS achieves the fastest convergence in comparison to all baseline methods within the provided time and memory constraints.

Table~\ref{tab:loss_further} provides results for our considered methods on the Poisson equation in $100d$ and $500d$ as proposed by~\citet{weinan2018deep}.  We demonstrate that our NWoS method achieves lower relative $L^2$-error than the baselines. However, we discover empirically that, for this benchmark, a simple projection to the boundary achieves the highest accuracy. This can be motivated by the smoothness of the solution and the fact that uniformly distributed evaluation samples concentrate at the boundary in high dimensions.

\begin{table}[t!]
\vskip -6pt
\centering
\caption{Relative $L^2$-error (and standard deviations over $5$ independent runs) of our considered methods, estimated using MC integration on $10^6$ uniformly distributed (unseen) points in $\Omega$.} 
\vspace{0.5em}
\resizebox{\linewidth}{!}{\begin{tabular}{lrr}
\toprule
Method & \multicolumn{2}{c}{Problem} \\ 
 &        Poisson ($100d$)  & Poisson ($500d$)  \\ \midrule
PINN & $1.49e^{-3}\pm3.21e^{-5}$ & $2.42^{-2}\pm6.06e^{-4}$ \\ 
Deep Ritz & $1.77e^{-2}\pm1.94e^{-4}$ & $9.92e^{-3}\pm2.56e^{-5}$\\ 
Diffusion loss & $6.71e{-4}\pm1.31e^{-5}$ &$9.47e^{-3}\pm3.81e^{-5}$\\ 
Projection & $\mathbf{2.92e^{-4}\pm5.17e^{-7}}$ &$\mathbf{1.19e^{-5}\pm1.67e^{-8}}$ \\ 
\textbf{NWoS (ours)} & $6.22e^{-4}\pm1.18e^{-5}$ & $9.14^{-3}\pm6.31e^{-5}$  \\ 
\bottomrule
\end{tabular}}
\label{tab:loss_further}
\end{table}

\end{document}